\pdfoutput=1

\documentclass[11pt]{article}

\usepackage{algorithm}
\usepackage{algpseudocode}

\usepackage[preprint]{acl}

\usepackage{times}
\usepackage{latexsym}

\usepackage{amsmath}
\usepackage{mathtools}
\usepackage{booktabs} 
\usepackage{amssymb} 
\usepackage{enumitem}

\usepackage[T1]{fontenc}

\usepackage[utf8]{inputenc}

\usepackage{microtype}

\usepackage{inconsolata}

\usepackage{graphicx}

\usepackage{multirow}

\usepackage{tabularx}
\usepackage{subfigure}

\usepackage{makecell}

\usepackage{xcolor}  

\title{TrendFact: A Benchmark Towards Hotspot Perception in Automatic Fact-Checking}

\author{Xiaocheng Zhang\textsuperscript{\rm 1}\textsuperscript{*},
    Xi Wang\textsuperscript{\rm 2}\thanks{Equal contribution},
    Yifei Lu\textsuperscript{\rm 3},
    Jianing Wang\textsuperscript{\rm 4},
    Zhuangzhuang Ye\textsuperscript{\rm 1}\\
    {\bf Mengjiao Bao\textsuperscript{\rm 5}},
    {\bf Peng Yan\textsuperscript{\rm 6}}\thanks{Corresponding Author},
    {\bf Xiaohong Su\textsuperscript{\rm 1}} \\
        \textsuperscript{\rm 1}Harbin Institute of Technology
        \textsuperscript{\rm 2}National University of Defense Technology\\
        \textsuperscript{\rm 3}Northeastern University
        \textsuperscript{\rm 4}East China Normal University
        \textsuperscript{\rm 5}Beihang University 
        \textsuperscript{\rm 6}Tsinghua University \\
        \texttt{zxcheng123cc@163.com},
        \texttt{wx\_23ndt@nudt.edu.cn}
        }

\begin{document}

\maketitle

\begin{abstract}
With the surge of online misinformation, Large Language Models (LLMs) and Reasoning Large Language Models (RLMs) serving as Automatic Fact-Checking (AFC) systems have emerged as a prominent paradigm for reliable, explainable verification. However, our empirical study reveals that this paradigm faces a critical risk asymmetry challenge when deployed in the real world under resource-constrained environments. While Hotspot Perception Ability (HPA), the capacity to dynamically allocate reasoning resources based on social impact, is essential to mitigate this risk, existing benchmarks lack the social metadata and evaluation framework to meet this urgent evaluation needs, thereby hindering the advancement of these AFC systems. To bridge this gap, we introduce TrendFact, the first benchmark capable of evaluating HPA and three fact-checking tasks. It consists of 7,643 curated samples sourced from trending platforms and professional datasets, with an evidence library containing 366,634 entries. To enable HPA assessment, we propose two novel metrics: the Explanation Consistency Score (ECS) to evaluate the reliability of verification reasoning, and the Hotspot Claim Perception Index (HCPI) to quantify the overall HPA of AFC systems. Extensive experiments demonstrate that existing AFC systems exhibit limited performance on TrendFact. Furthermore, our proposed FactISR framework effectively enhances HPA and computational efficiency for RLMs-served AFC systems. 

\end{abstract}

\section{Introduction}
\begin{figure}[!ht]
\centering
\includegraphics[width=1\linewidth]{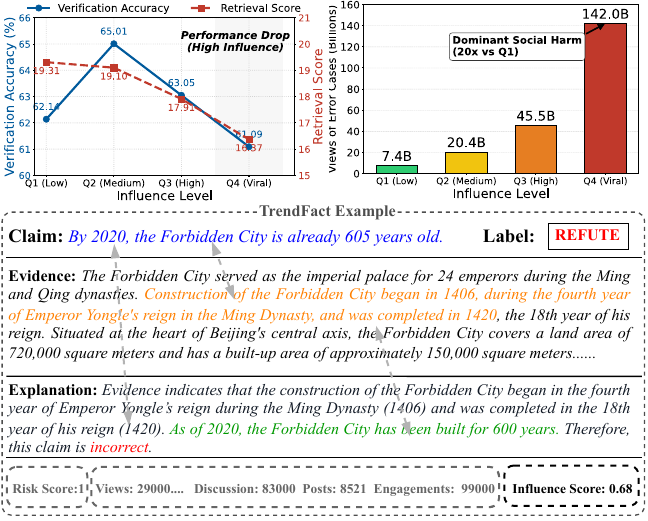} 
\caption{An illustration of the risk asymmetry challenge in LLMs/RLMs-served AFC systems and a representative sample from the TrendFact benchmark.}
\label{introduction_fig}
\end{figure}

The proliferation of counterfeit claims poses significant social risks, including mass panic, social destabilization, and even armed conflicts, as exemplified by the COVID-19 infodemic \cite{van2020inoculating, aondover2024propagation}. This critical challenge has driven substantial research efforts in Automatic Fact-Checking (AFC) systems. With the rapid evolution of Large Language Models (LLMs) and Reasoning Language Models (RLMs) \cite{atanasova2024generating,rani2023factify}, generating natural language explanations to enhance the transparency and trustworthiness of AFC has emerged as a research hotspot \cite{wang2023explainable,bilal2024generating,kao2024we}. Unlike traditional black-box classifiers, this paradigm not only provides verification labels but also leverages structured reasoning processes as evidence, significantly bolstering user trust in the verification results.

However, our empirical study reveals a severe challenge, the risk asymmetry, when this checking paradigm is employed in the real world under resource-constrained environments \cite{vosoughi2018spread, chen2023frugalgpt,zhang2026stable}. As illustrated in the upper-left of Figure \ref{introduction_fig}, high-influence claims show heightened verification difficulty under identical resource settings. Furthermore, the upper-right of Figure \ref{introduction_fig} shows that incorrect verification of such hotspots can trigger social consequences up to 20 times more severe than those of low ones. To mitigate these risks, an ideal AFC system should possess \textbf{Hotspot Perception Ability (HPA)}, the capacity to dynamically schedule reasoning resources based on claim’s social impact. For instance, allocating more evidence and "thinking steps" to high-influence claims to minimize misjudgment, while streamlining resources for low-influence claims to enhance overall efficiency.

Despite the necessity of HPA, current benchmarks remain inadequate to guide the design and iteration of such systems. A prerequisite for evaluating HPA is a benchmark that provides not only foundational attributes (e.g., labels, evidence) but also data reflecting real-world dissemination dynamics, such as social media metadata (e.g., views, posts). Moreover, as reasoning faithfulness is the cornerstone of LLMs/RLMs-served AFC, high-quality textual explanations are indispensable for reliable assessment. However, existing benchmarks primarily focus on label accuracy, neglecting the critical social influence and reasoning dimensions required for a comprehensive HPA evaluation.

To bridge this gap, we construct \textbf{TrendFact}, the first benchmark capable of evaluating HPA and covering three fact-checking tasks. It comprises 7,643 samples with an evidence library of 366,634 pieces, built through a rigorous logical progression. Specifically, TrendFact integrates four types of social metadata, views, discussions, engagements, and posts, combined with a GPT-evaluated risk score to quantify the Influence Score for each claim. Additionally, TrendFact provides human-annotated explanations and designs the \textbf{Explanation Consistency Score (ECS)} to assess the reliability of reasoning reliability. Finally, we propose a comprehensive HPA metric, the \textbf{Hotspot Claim Perception Index (HCPI)}, which fuses the influence score with ECS to quantitatively measure comprehensive HPA performance across varying social impacts on TrendFact. Extensive experiments show that existing fact-checking methods and LLMs/RLMs-served systems all exhibit limited performance on TrendFact.

Furthermore, to address the deficiencies of RLMs-served systems revealed by TrendFact, we propose \textbf{FactISR}, an influence-aware enhancement framework that incorporates the reasoning process with dynamic resource scheduling capabilities. FactISR synergizes dynamic evidence augmentation for on-demand iterative retrieval with influence-driven self-reflection for adaptive reasoning depth. This ensures that deeper cognitive effort and sufficient evidence are prioritized for high-stakes claims. Experimental results confirm that FactISR achieves significant performance gains while improving overall computational efficiency.

In summary, our contributions are as follows:
\begin{itemize}
    \item Our empirical analysis reveals the risk asymmetry challenge in LLMs/RLMs-served AFC and identifies the critical absence of HPA evaluation in existing benchmarks.
    \item We construct TrendFact, the first benchmark capable of HPA evaluation with comprehensive coverage of fact-checking tasks. It enables the quantitative assessment of HPA through the HCPI metric, which integrates the ECS and the influence score to ensure both reasoning reliability and social impact.
    \item We propose FactISR, an influence-aware framework to enhance RLMs-served AFC through dynamic evidence augmentation and self-reflection tailored to claim impact.
    \item Extensive experiments demonstrate the limitations of current fact-checking methods on TrendFact and verify that FactISR improves both HPA performance and resource efficiency of RLMs-served AFC\footnote{Code and data are available at \href{https://github.com/zxc123cc/TrendFact}{https://github.com/{\\} zxc123cc/TrendFact}}.
\end{itemize}

\begin{table*}[!t]
\centering
\renewcommand{\arraystretch}{1.05} 
\resizebox{1\textwidth}{!}{
\begin{tabular}{lcccccccc}
\toprule
\multirow{3}{*}{\textbf{Dataset}} 
& \multirow{3}{*}{\textbf{\#Claims}}  
& \multirow{3}{*}{\textbf{Source}}  
& \multirow{3}{*}{\textbf{Language}}  
& \multirow{3}{*}{\raisebox{-1.3ex}{\makecell[c]{\textbf{Explanation}\\\textbf{Contains}}}} 
& \multirow{3}{*}{\textbf{HPA}}
& \multicolumn{3}{c}{\textbf{Task}} \\
\cmidrule{7-9}
            &                               &                &       &   &   & \textbf{Evidence}    & \textbf{Claim}  & \textbf{Explanation}        \\
            &                                 &                &         &  &  & \textbf{Retrieval} & \textbf{Verification}  & \textbf{Generation} \\
\midrule
\textbf{Synthetic Claims} & & & & & & & \\ 
FEVEROUS\cite{aly2021feverous}         &  $87,026$    & WP & English & $\times$ & $\times$ & Yes & Yes & No \\
CHEF\cite{hu2022chef}          &  $10,000$  & FCS & Chinese & $\times$ & $\times$ & Yes & Yes & No \\
Hover\cite{jiang2020hover}            &  $26,171$  & WP & English & $\times$ & $\times$ & Yes & Yes & No \\
CFEVER\cite{lin2024cfever}           &  $30,012$  & WP & Chinese & $\times$ & $\times$ & Yes & Yes & No \\
STATPROPS\cite{thorne2017extensible}        &  $4,225$  & FB & English & $\times$ & $\times$ & Yes & Yes & No \\
\midrule
\textbf{Fact-checker Claims} & & & & & & & \\ 
CLAIMDECOMP\cite{chen2022generating}      &  $1,250$  & Politifact & English & $\times$ & $\times$ & Yes & Yes & No \\
DeClarE\cite{popat2018declare}          &  $13,525$  & FCS & English & $\times$ & $\times$ & Yes & Yes & No \\
X-Fact\cite{gupta2021x}           &  $1,800$  & FCS & Multi & $\times$ & $\times$ & Yes & Yes & No \\
AVeriTeC\cite{schlichtkrull2024averitec}         &  $4,568$  & FCS & English & $\times$ & $\times$ & Yes & Yes & No \\
FlawCheck\cite{kao2024we} &  $30,349$  & FCS & English & $\checkmark$ & $\times$ & Yes & Yes & Yes \\
QUANTEMP\cite{venktesh2024quantemp}         &  $30,012$  & FCS & English & $\times$ & $\times$ & Yes & Yes & No \\
\midrule
TrendFact         &  $7,643$  & TP & Chinese & $\checkmark$ & $\checkmark$ & Yes & Yes & Yes \\
\bottomrule
\end{tabular}
}
\caption{Comparison of TrendFact with other fact-checking datasets. WP refers to Wikipedia, FCS refers to fact-checking websites, FB refers to FreeBase, and TP refers to Trending Platforms. By 'HPA', we refer to whether the dataset supports the evaluation of a fact-checking system's hotspot perception ability.}
\label{benchmarks}
\end{table*}

\section{Related Work}
\paragraph{Fact-checking Benchmarks}
Existing fact-checking benchmarks are generally divided into two categories: one based on Wikipedia data, such as Hover \cite{jiang2020hover} and FEVER \cite{thorne2018fever}, and another using knowledge bases from fact-checking websites, such as CLAIMDECOMP \cite{chen2022generating} and QUANTEMP \cite{venktesh2024quantemp} (as shown in table \ref{benchmarks}). These benchmarks primarily focus on fact verification and evidence retrieval tasks, often neglecting explanation generation evaluation. With the growing role of LLMs and RLMs in generating explanations, evaluating their reliability is crucial. Moreover, it is also a pressing issue in the field to evaluate the Hotspot Perception Ability (HPA) of Automatic Fact-Checking (AFC) systems \cite{solovev2022moral, sehat2024misinformation}, which refers to the scheduling of reasoning resources based on a claim’s social impact. Therefore, we construct TrendFact benchmark, which evaluates both explanation generation reliability and hotspot perception, is essential for comprehensive fact-checking evaluation.

\paragraph{Automatic Fact-checking}
Research on Automatic Fact-Checking primarily falls into two categories: fact verification and explanation generation. Fact verification focuses on timely claim evaluation and has been widely explored in contexts such as Wikipedia articles, table-based data, and QA dialogues. With the rise of LLMs and RLMs, methods like PROGRAMFC \cite{pan2023fact} generate executable programs to support step-by-step verification. Explanation generation aims to produce interpretable outputs, but most work treats it as an intermediate step rather than a core objective. Few studies explore using natural language to convey both claim veracity and reasoning, which is critical for model interpretability and user understanding. For instance, \citet{he2023reinforcement} generates counter-misinformation responses to correct false claims. Additionally, many methods overlook the HPA, which is essential for improving fact-checking systems' ability to solve different influential claims. Incorporating HPA, along with the interplay between verification and explanation, can enhance transparency, trust, and the solving capability of high influence claims in the real world.

\begin{figure*}[!ht]
\centering
\includegraphics[width=1\linewidth]{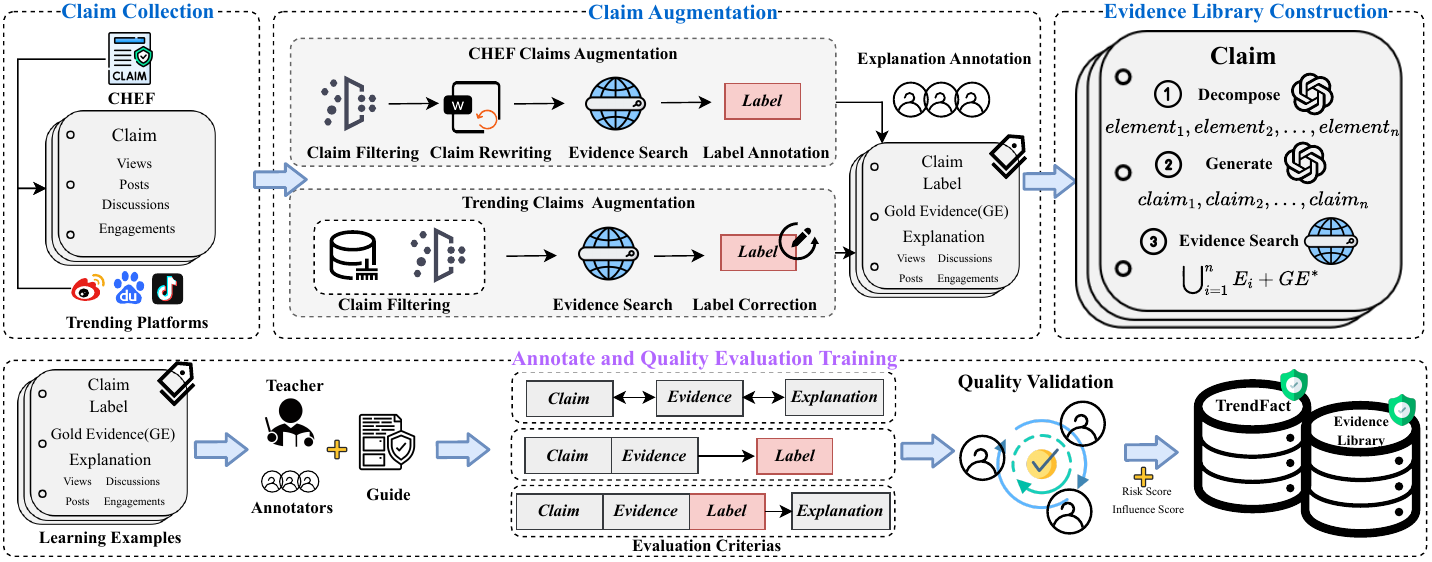} 
\caption{Overview of TrendFact. The overall construction process of TrendFact includes claim collection, filtering, augmentation, evidence library construction, and a multi-stage sample review process.}
\label{framework}
\end{figure*}

\section{The TrendFact Benchmark}



Aforementioned empirical analysis (see Figure \ref{introduction_fig}) demonstrates that existing LLMs/RLMs-served Automatic Fact-Checking (AFC) systems face a critical risk asymmetry challenge. We first attribute this vulnerability to the Hotspot Perception Ability (HPA), the capacity to dynamically modulate reasoning resources based on claim influence. Then, the advancement of HPA of AFC systems is also hindered by the limitations of existing benchmarks, which focus solely on static accuracy and lack the social influence metadata essential for HPA evaluation. Therefore, we propose the TrendFact benchmark to bridge this critical gap.

In this section, we detail the dataset construction and evaluation metrics formulation of our proposed TrendFact, the first benchmark that incorporates real-world hotspot indicators to establish influence perception as a core evaluation dimension. 

\subsection{Dataset Construction}
In this section, we detail the dataset construction process of TrendFact. As illustrated in Figure \ref{framework}, the process comprises three phases: collecting claims alongside critical hotspot indicators from diverse sources, augmenting data to ensure verifiability, and constructing a challenging evidence library.

\subsubsection{Data Attributes}
\label{sec_attributes}
TrendFact covers five primary domains: public health, science, society, politics, and culture. As shown in Figure \ref{introduction_fig}, each sample is annotated with essential fact-checking attributes including claim, label, gold evidence, textual explanation, risk score and unique hotspot metadata. Specifically, metadata includes four hotspot indicators (views, discussions, posts, and engagements) and a derived influence score, which serves as a key attribute to calculate the subsequent Hotspot Perception evaluation metrics. Based on gold evidence density, samples are categorized into single-evidence (85\%) and multi-evidence (15\%). Detailed definitions are provided in Appendix \ref{app:detial_data_attributes}.

\subsubsection{Claim Collection}
In contrast to previous datasets that primarily relied on sources with limited timeliness and popularity (e.g., Wikipedia), TrendFact collects claims from two complementary sources to capture real-world dissemination dynamics.
\noindent\textbf{Source One: Trending Platforms.} We collect claims from Weibo, DouYin, and Baidu, which provide large-scale, dynamic factual statements accompanied by hotspot metadata. The hotspot metadata is the four indicators detailed in section \ref{sec_attributes}. In total, we obtain approximately 500,000 raw claims from these platforms between 2020 and 2024.
\noindent\textbf{Source Two: Existing Datasets.} We further collect claims from existing fact-checking datasets to broaden verification scenarios beyond trending platforms. Although such datasets lack hotspot indicators, they offer valuable complex verification logic. In particular, we incorporate claims from the CHEF dataset, which aggregates data from multiple fact-checking websites, as a representative supplement.

\subsubsection{Data Augmentation}
The collected raw claims exhibit distinct limitations when applied to fact-checking tasks: trending claims often lack verifiable structures and essential attributes (e.g., labels, evidence), while claims from existing datasets frequently contain judgmental phrasing and lack textual explanations. To address these, we design tailored augmentation pipelines.
\paragraph{Trending Claims.}
We first  apply an LLM voting mechanism to filter out approximately 90\% of noise, specifically targeting entertainment trivia, interrogative clickbait, and unstructured statements lacking assessable factual content (details in Appendix \ref{filtering_prompt}). Then, our annotation team removes sensitive or overlapping content, resulting in 6,512 claims. Since the remaining raw claims are often unstructured, annotators transform them into verifiable fact-checking claims following a specific rewriting guideline focused on resolving ambiguity and supplementing missing context (e.g., temporal or domain constraints). Comprehensive details of the rewriting factors are available in Appendix \ref{app:rewrite_factor}.
\paragraph{CHEF Claims.}
We first filter samples for factual accuracy and sensitivity, then use the same LLM voting mechanism to select claims that present significant verification challenges. Each selected claim is further annotated with a detailed explanation by our team, ensuring the same explanatory standard applied to trending claims. This process results in 1,131 enhanced claims.

After augmentation, we merge two sets to construct TrendFact dataset, consisting of 7,643 samples. To ensure reliability and quality of annotations, we implement rigorous control mechanisms. Annotators underwent comprehensive training to ensure accuracy. The training process covers examples of claim rewriting, emphasizing the importance of maintaining semantic integrity while transforming  claim into verifiable, fact-checkable statement. As shown in Figure \ref{framework}, all rewritten and annotated outputs are validated through a three-level standardized human evaluation criteria, ensuring semantic integrity and label consistency across the benchmark (details in Appendix \ref{app:human_evaluation_criteria}).

\subsubsection{Evidence Library Construction}
As illustrated in Figure \ref{framework}, the evidence library construction of the TrendFact follows a three-stage process: decomposition, generation, and retrieval. Specifically, we first employ an LLM to extract three or more key elements from each claim. We generate new claims for elements and retrieve supporting evidence from websites, which is then combined with the gold evidence from the initial TrendFact dataset to form a comprehensive evidence library containing 366,634 entries. More detailed descriptions are provided in Appendix \ref{ELC:more_details}.

\subsection{Metric Formulation: Quantifying Hotspot Perception Ability (HPA)}
It is essential that a reliable AFC system provides trustworthy reasoning for its judgments, ensuring its verification results are deduced from evidence rather than merely predicted. Therefore, we formulate our evaluation framework by integrating verification reliability evaluation metric, ECS, into the final HPA assessment metric, HCPI.
\paragraph{Verification Reliability: Explanation Consistency Score (ECS).} 

Generating a correct label based on flawed reasoning (i.e., hallucination) fundamentally undermines the credibility of an AFC system. To quantify this reasoning fidelity, we introduce ECS as a system-level evaluation metric—designed to assess benchmark-wide consistency and interpretability, rather than serving as a user-facing trustworthiness signal. Specifically, ECS assesses the consistency between the generated explanation and the gold standard, relative to the predicted label. As detailed in Table \ref{ECS}, we establish a five-level scoring rubric in which five discrete consistency levels are normalized and mapped onto the interval 
$[0.2, 1.0]$. It assigns graded reliability weights to verification results, significantly attenuating the contribution of ``correct but hallucinated'' samples (e.g., Category T-CD) while affirming logically consistent predictions with high confidence scores (e.g., Category T-FC).
To verify the reliability of our LLM-as-a-Judge evaluator, we conduct a human expert validation study, with detailed results provided in Appendix \ref{sec:LaaJ}.

\begin{figure*}[!ht]
\centering
\includegraphics[width=1\linewidth]{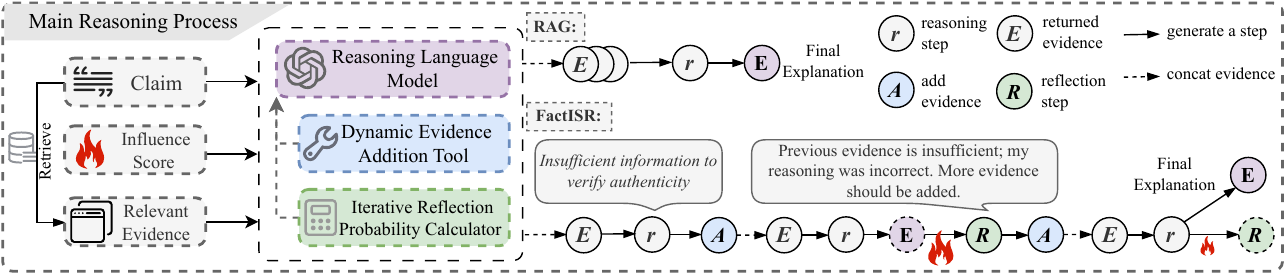} 
\caption{Overview of FactISR. The bottom-right section shows the reasoning process of FactISR, where "flame" represents the reflection probability calculated based on influence score, which continuously decays with the number of iterative reflections. The reasoning process terminates when the maximum number of reflections is exceeded or no reflection step is sampled.}
\label{FactISR}
\end{figure*}

\begin{table}[t]
\centering
\resizebox{1\linewidth}{!}{
\begin{tabular}{ccc|c}
\toprule
\textbf{Category} & \textbf{Label Acc} & \textbf{Explain Cons} & \textbf{Score} \\ \midrule
F-D & F & Full Discrepancy & 0.2 \\ 
F-C & F & Consistency & 0.4 \\ 
T-CD & T & Content Divergence & 0.6 \\ 
T-PC & T & Partial Consistency & 0.8 \\ 
T-FC & T & Full Consistency & 1.0 \\ \bottomrule
\end{tabular}}
\caption{Definition of ECS. Label Acc indicates the accuracy of the fact verification, and Explain Cons represents the consistency of the explanation generated by the system and the gold one that evaluated by LLM .}
\label{ECS}
\end{table}

\paragraph{Integration: Hotspot Claim Perception Index (HCPI).}
We formally define the HCPI metric to quantify an AFC system's HPA by dynamically weighting each claim based on its influence score, verification label, and the reliability coefficient, ECS. 
First, we calculate an influence score $s_i$ for each claim $c_i$ to capture its social impact. This score integrates the risk level ($r_i$) evaluated by GPT and four hotspot indicators, views $v$, discussions $d$, engagements $e$, and posts $p$, formulated as:
\begin{equation}
\label{eq:influence_score}
s_i = r_i \cdot \sum_{x \in {v, d, e, p}} w_x \cdot \log(1+x_i)
\end{equation}
where $w_x$ denotes the weight for each indicator derived from the statistical distribution of the raw hotspot data, and the logarithmic term smooths the heavy-tailed distribution of social media metrics. A sensitivity analysis with $\pm20\%$ perturbations on all $w_x$ shows negligible impact on HCPI scores and fully consistent model rankings, confirming the robustness of our metric.
Based on this score, we then define the HCPI metric as the normalized influence-weighted score of $N$ claims in TrendFact:
\begin{equation}
\label{eq:hcpi_main}
\text{HCPI} = \frac{\sum_{i=1}^{N} \mathcal{S}(c_i, \hat{y}_i)}{\sum_{i=1}^{N} s_i},
\end{equation}
where the numerator aggregates the AFC system's performance, while the denominator represents the total social influence across the benchmark.

The core of HCPI lies in the scoring function $\mathcal{S}(\cdot)$, which integrates the $\text{ECS}_i$ and applies asymmetric penalties based on error types. Let $y_i$ denote the ground truth label and $\hat{y}_i$ represent the system's predicted label. The scoring function is defined as:
\begin{equation}
\label{eq:hcpi_logic}
\small
\mathcal{S}(c_i, \hat{y}_i) = 
\begin{cases} 
    s_i \cdot \text{ECS}_i & \text{if } \hat{y}_i = y_i \\
    -2 \cdot s_i & \text{if } y_i = \textsc{Sup} \land \hat{y}_i = \textsc{Ref} \\
    -1 \cdot s_i & \text{if } y_i = \textsc{Nei} \land \hat{y}_i = \textsc{Ref} \\
    0 & \text{otherwise}
\end{cases}
\end{equation}
\noindent The design rationale for this function is as follows:
\begin{itemize}
\item \textbf{Reliability-Weighted Score ($s_i \cdot \text{ECS}_i$):} For samples with correct verification labels, we multiply influence score by ECS-served reliability coefficient, to weight the final outcome. This ensures that the evaluation accounts for both trustworthy reasoning and influence-aware essential for AFC systems.
\item \textbf{Critical and Conservative Penalty ($-2s_i$ and $-1s_i$):} We impose varying penalties for incorrect predictions labeled as \textsc{Refute}. Specifically, a severe penalty is applied for falsely debunking a true claim ($y_i=\textsc{Sup}, \hat{y}_i=\textsc{Ref}$), as labeling a true event as a rumor causes fatal damage to credibility. Similarly, a moderate penalty is used for fabricating a debunking verdict without sufficient evidence ($y_i=\textsc{Nei}, \hat{y}_i=\textsc{Ref}$), thereby discouraging unfounded judgments.
\end{itemize}

\section{FactISR}
In this section, we detail our proposed method, FactISR, a reasoning framework designed to enhance the fact-checking capabilities and Hotspot Perception Ability (HPA) of RLMs. As illustrated in Figure \ref{FactISR}, it consists of two key components: the Dynamic Evidence Augmentation (DEA) and the Iterative Self-Reflection (ISR). Unlike traditional RAG, which loads evidences at once for reasoning, FactISR dynamically leverages DEA to supplement evidence throughout the reasoning process. And based on the claim's influence score and the intermediate reasoning results, ISR is selectively activated by RLM to further refine this process and achieve more accurate outcomes.

\paragraph{Dynamic Evidence Augmentation (DEA)} Traditional RAG process loads all evidence at once, which can lead to issues such as insufficient, redundant, or irrelevant evidence, hindering accurate judgment. To address these, FactISR’s DEA module dynamically retrieves and adds relevant evidence during the reasoning process based on ongoing analysis of RLMs. This continuous addition of evidence guarantees more refined evidence and produces better fact-checking performance.

\paragraph{Iterative Self-Reflection (ISR)}
Traditional RAG evaluates all claims equally and lacks the capability to adjust its reasoning budget based on a claim's popularity. Existing reflection mechanisms are typically reactive, triggering only when a model detects its own uncertainty—a model-centric paradigm that risks overlooking the social impact of high-influence claims. This limitation can lead to redundant inference for low-influence claims and insufficient reasoning for high-influence ones, ultimately detracting from HPA performance. To address these, we introduce the Iterative Self-Reflection module, which shifts the reflection trigger from internal model uncertainty to external social risk. Rather than relying on the model's self-assessed confidence, ISR proactively initiates deeper reasoning when a claim is identified as having high social impact, ensuring that high-stakes claims receive rigorous verification while computational resources are allocated efficiently. This risk-centric reasoning paradigm is particularly essential in real-world deployment scenarios where resources are finite.

Specifically, if a claim has a high influence score, the probability of reflection reasoning increases. Moreover, as the reasoning process advances, the probability gradually decreases, effectively preventing excessive reflection.  The formal definition of the reflection probability is as follows:
\begin{equation}
P_{i}^n = ( k \cdot {\frac{s_i - s_{\text{min}}}{s_{\text{max}} - s_{\text{min}} + \epsilon}} ) \cdot \gamma^{n-1}
\end{equation}

Where \( P_{i}^n \) represents the probability of the \( n \)-th reflection for the \( i \)-th sample, while \( s_{\text{min}} \) and \( s_{\text{max}} \) denote the minimum and maximum scores among all claims, respectively.
A small constant \( \epsilon \) is included to prevent division by zero, typically set to \( 1 \times 10^{-8} \). The hyperparameter \( k  \space \in [0,1]\) is crucial for flexibly controlling the overall reflection intensity across samples, with a default value of 1, and \(\gamma\space \in [0,1]\) is the decay factor.
where the superscript \( n \) in \( \gamma^n \) indicates exponentiation.

\begin{table}[t]
\centering
\small 
\renewcommand{\arraystretch}{1.2} 
\resizebox{\columnwidth}{!}{
\begin{tabular}{l|cccc}
\toprule
Methods            & R@1   & R@2   & R@3   & R@5  \\ \midrule
BM25 -w/o \textit{date}   & 12.08  & 20.46 & 25.98 & 33.70  \\
BM25               & 12.99 & 20.76 & 26.65 & 34.91 \\
text-emb-ada-002 & 12.10 & 21.44 & 26.88 & 35.42 \\
bge-m3(dense)             & \textbf{14.10} & \textbf{22.76} & \textbf{29.63} & \textbf{39.97} \\
 \bottomrule
\end{tabular}}
\caption{Experimental Results on Evidence Retrieval.}
\label{tab:retrieval_results}
\end{table}

\section{Experiment}
\subsection{Setup}
\paragraph{Metrics}
For evidence retrieval task, we choose R@k, where k=1,2,3,5.
For verification task, we choose F1-macro, Precision, Recall, and Accuracy. 
For explanation generation task, in addition to ECS, we also employ BLEU-4, ROUGE-(1, 2, L), and BERTScore. 
For assessing the HPA of fact-checking systems, we employ HCPI.

\paragraph{Baselines}

In this work, we choose the following types of methods as baseline, including RLMs, LLMs, and existing fact-checking methods. For RLMs, we select the most advanced QwQ-32B, QwQ-32B-Preview \cite{qwen2}, Qwen3-32B (Think) \cite{yang2025qwen3}, and DeepSeek-R1-0528 \cite{guo2025deepseek}. For LLMs, we choose GPT-4.1 \cite{hurst2024gpt}, DeepSeek-v3 \cite{liu2024deepseek}, and Qwen2.5-72B-Instruct \cite{qwen2}, Qwen3-32B (No Think). For fact-checking methods, we select PROGRAMFC \cite{pan2023fact} and CLAIMDECOMP \cite{chen2022generating}. For retrieval methods, we choose the following advanced methods, including BM-25 (without date), BM-25, OpenAI’s text-embedding-ada-002, and bge-m3 (dense) \cite{chen2024bge}.

\paragraph{Experimental Settings}
The detailed experiment settings are provided in the Appendix \ref{Experimental_Settings:more_details}.

\subsection{Main Results}

\paragraph{Evidence Retrieval Results}
We evaluate the selected retrieval methods based on their ability to retrieve the target gold evidence from the TrendFact evidence library. As shown in Table \ref{tab:retrieval_results}, the best performing method, bge-m3, achieves a relatively low performance with an R@5 less than 40\%. This suggests that our unique evidence library construction effectively gathers challenging evidence that distinguishes original gold evidence, thereby increasing the difficulty.

\begin{table}[t]
\centering
\small 
\renewcommand{\arraystretch}{1.2} 
\resizebox{\columnwidth}{!}{
\begin{tabular}{l|cccc}
\toprule
Methods                 & F1    & P & R  & Acc\\ \midrule
PROGRAM-FC            & 40.46 & 41.66 & 43.30  & 43.17\\
CLAIMDECOMP           & 42.53 & 44.18 & 44.90  & 45.28\\ \midrule
QwQ-32B-Preview       & 49.52 & 52.32     & 53.48  & 55.29 \\
Qwen2.5-72B-instruct  & 47.10 & 53.51    & 52.51  & 55.96 \\
Qwen3-32B(\textit{No think})   & 50.06 & 53.00     & 53.92 & 58.84 \\
DeepSeek-V3-0324      & 52.10 & 55.19     & 55.94  & 60.67 \\
GPT-4.1               & 52.45 & 56.23     & 55.88 & 61.29 \\
Qwen3-32B(\textit{Think})  & 58.14 & 58.05     & 59.82 & 66.09 \\
QwQ-32B               & 58.58 & 58.89     & 60.09  & 68.61 \\
DeepSeek-R1-0528      & 58.89 & 59.15     & 60.58 & 68.44 \\
\midrule
FactISR(\textit{Qwen3-32B})     &58.68  & 58.55     & 60.84 &67.49 \\ 
FactISR(\textit{QwQ-32B})      & \textbf{61.17} & \textbf{61.04}     & \textbf{63.37} & \textbf{69.70} \\
\bottomrule
\end{tabular}}
\caption{Comparison of FactISR with other baselines on fact verification task.}
\label{tab:compare_baseline_fact_verification}
\end{table}

\begin{table*}[]
\small 
\centering
\renewcommand{\arraystretch}{1.2} 
\resizebox{\textwidth}{!}{
\begin{tabular}{l|ccccccc}
\toprule
Methods              & HCPI & ECS & BLEU-4 & BERTScore & ROUGE-1 & ROUGE-2 & ROUGE-L  \\ \midrule
QwQ-32B-Preview    &  0.4820 &  0.7843  & 0.1573     & 0.7525  &  0.4699  & 0.2781  & 0.4048 \\ 
Qwen2.5-72B-instruct    &  0.5061 &  0.7207  & \textbf{0.2632}     & 0.8015  &  0.5637  & \textbf{0.3712}  & 0.5080 \\ 
Qwen3-32B(\textit{No think}) &  0.5001 &  0.7574  & 0.2491     & \textbf{0.8049}  &  \textbf{0.5649}  & 0.3658  & \textbf{0.5106} \\  
DeepSeek-V3-0324          &  0.5427 &  0.7674  & 0.2360     &  0.7918  &  0.5423  & 0.3427  & 0.4887 \\
GPT-4.1               & 0.5488  & 0.7726   & 0.2214     & 0.7735  & 0.5380   & 0.3341 & 0.4736 \\
Qwen3-32B(\textit{Think}) &  0.5575 & 0.8332  & 0.2181     & 0.7836  &  0.5188  &  0.3138  & 0.4603 \\
QwQ-32B                  &  0.5621 &  0.8531  & 0.1940     & 0.7712  &  0.4896  & 0.2864  & 0.4247 \\
DeepSeek-R1-0528          &  0.5681 &  0.8399  & 0.2041     & 0.7764  &  0.5016  & 0.2954  & 0.4383 \\ \midrule
FactISR(\textit{Qwen3-32B})    & 0.5926  & 0.8426   & 0.2281    & 0.7912  & 0.5365   & 0.3334  & 0.4833 \\
FactISR(\textit{QwQ-32B})      &  \textbf{0.6010} & \textbf{0.8610}   & 0.2175     & 0.7835  & 0.5156   & 0.3107  & 0.4592 \\
\bottomrule
\end{tabular}}
\caption{Comparison of FactISR with Other Baselines on Explanation Generation.}
\label{tab:compare_baseline_explanation_generation}
\end{table*}

\paragraph{Fact Verification Results}

We conduct a comprehensive evaluation of selected baselines on the verification task of TrendFact, with the results presented in Table \ref{tab:compare_baseline_fact_verification}. The key findings are as follows: First, traditional fact-checking methods perform the worst, with all verification scores falling below 50\%. Second, both LLMs and RLMs consistently perform better than 50\%. Notably, RLMs outperform LLMs, as they are better equipped to handle the complex reasoning required by many TrendFact samples. However, even the best-performing DeepSeek-R1-0528 fails to achieve an overall verification score above 60\%, highlighting the significant challenge posed by the high-quality and reasoning-intensive nature of TrendFact. Moreover, FactISR contributes to improved verification performance for RLMs. For instance, it enhances the performance of the RLM QwQ-32B, enabling it to achieve an overall verification F1 score of 61, surpassing DeepSeek-R1-0528.

\paragraph{Explanation Generation Results}
We evaluate both LLMs and RLMs on the explanation generation task in TrendFact, with the results presented in Table \ref{tab:compare_baseline_explanation_generation}. Our findings are as follows:
First, RLMs generally produce lower-quality explanations compared to LLMs on surface-level metrics such as BLEU and ROUGE. However, this gap actually highlights the limitations of these metrics—they measure surface-level lexical similarity rather than the validity of the reasoning process, which is precisely why we introduced ECS. We attribute this phenomenon to the design philosophy of RLMs: their training places greater emphasis on optimizing reasoning logic, at the cost of output fluency and linguistic style. This observation is consistent with the findings reported in several RLM technical reports \cite{guo2025deepseek}, which similarly note that traditional metrics often fail to capture the true reasoning quality of RLMs, and explicitly highlight challenges such as poor readability and language mixing. Since FactISR is built upon an RLM and deliberately avoids heavy few-shot prompting with reference examples, its lower explanation quality compared to non-reasoning LLMs is expected.
Second, RLMs outperform LLMs in explanation consistency, as measured by ECS. This is expected, as ECS reflects the alignment between explanations and the model's internal reasoning process. Furthermore, FactISR enhances both the explanation quality and consistency of RLMs. For example, FactISR improves the performance of QwQ-32B by 1–3 percentage points across all explanation metrics, enabling it to approach the performance of LLMs in explanation quality. This demonstrates that, despite the inherent disadvantage of RLMs on surface-level text metrics, FactISR can effectively improve the explanatory expressiveness of RLMs without relying on extensive reference examples.

\begin{table}[t]
\centering
\small 
\renewcommand{\arraystretch}{1.2} 
\resizebox{\columnwidth}{!}{
\begin{tabular}{l|ccc}
\toprule
Methods              & F1   & ECS    & HCPI \\ \midrule
FactISR(\textit{QwQ-32B})      & \textbf{61.17} & \textbf{0.8610} & \textbf{0.6010}      \\ 
- w/o \textit{DEA} & 59.79  & 0.8501 & 0.5833      \\
- w/o \textit{ISR} & 58.88  & 0.8388  & 0.5675     \\
\midrule
FactISR(\textit{Qwen3-32B})      & \textbf{58.68} & \textbf{0.8426} & \textbf{0.5926}      \\
- w/o \textit{DEA} & 58.46 & 0.8372 & 0.5785      \\
- w/o \textit{ISR} &58.30  & 0.8390 & 0.5650      \\ \bottomrule
\end{tabular}}
\caption{Ablation Study for Evaluating Each Component of method FactISR.}
\label{tab:xiaorongshiyan}
\end{table}

\paragraph{HPA Assessment Results}
Since the verification results affect the calculation of HCPI, RLMs naturally exhibit a higher HCPI compared to LLMs. However, without the capability to adjust reasoning budgets according to claim influence, RLMs risk redundant inference on low-influence claims and insufficient reasoning on high-influence ones, ultimately undermining HPA performance. In contrast, FactISR effectively addresses this imbalance via the influence score-based ISR module combined with DEA, improving the RLM's HCPI score by up to nearly 4\%. Notably, FactISR's modest gains on standard metrics stem from our resource-constrained experimental setup (average of 3 evidence items per claim), under which standard metrics—by weighting all claims equally—inherently fail to capture the targeted robustness on high-impact events, which is precisely what HCPI is designed to quantify.

\begin{table}[t]
\centering
\small 
\renewcommand{\arraystretch}{1.2} 
\resizebox{\columnwidth}{!}{
\begin{tabular}{l|ccc}
\toprule
Methods              & F1   & Time    & Length \\ \midrule
QwQ-32B      & \textbf{58.89} & 0.5960 & 2664      \\
+ \textit{DEA} & 58.88 & \textbf{0.4594}\ (\scriptsize$\downarrow$22.92\%) & \textbf{1442}\ (\scriptsize$\downarrow$47.31\%)        \\
\midrule
Qwen3-32B      & 58.14 & 0.5515 & 2664      \\
+ \textit{DEA} & \textbf{58.30} & \textbf{0.4380}\ (\scriptsize$\downarrow$30.17\%) & \textbf{1316}\ (\scriptsize$\downarrow$50.62\%)     \\ \bottomrule
\end{tabular}}
\caption{Impact of DEA on Per-Sample Generation Time and Input Evidence Length.}
\label{tab:DEA_ablation}
\end{table}

\subsection{Ablation Study}

\paragraph{Component Effectiveness} We conduct experiments by individually removing one of the two modules from the fully integrated QwQ-32B and Qwen3-32B. We evaluate comprehensive performance using accuracy, ECS, and HCPI. The results are shown in Table \ref{tab:xiaorongshiyan}. It demonstrates that removing any single module leads to performance degradation across all tasks, confirming the effectiveness of all components in FactISR.

\paragraph{DEA Efficiency} 
We conduct experiments to analyze the efficiency gains of the DEA module. Unlike traditional RAG, which loads all evidence at once, DEA dynamically introduces evidence to mitigate redundancy and information overload, optimizing reasoning quality and efficiency rather than directly boosting verification accuracy. As shown in Table \ref{tab:DEA_ablation}, DEA maintains accuracy while achieving maximum reductions of 30\% and 50\% in reasoning time and length, respectively. Notably, the marginal performance drop in the ablation study (w/o DEA) is largely attributable to the per-claim evidence limit of 3 items imposed for experimental fairness, which compresses DEA's practical benefit space. These results demonstrate that DEA effectively mitigates reasoning inefficiencies caused by lengthy evidence, making it particularly valuable in resource-constrained settings.

\begin{figure}[t]
\centering  
\includegraphics[width=0.235\textwidth]{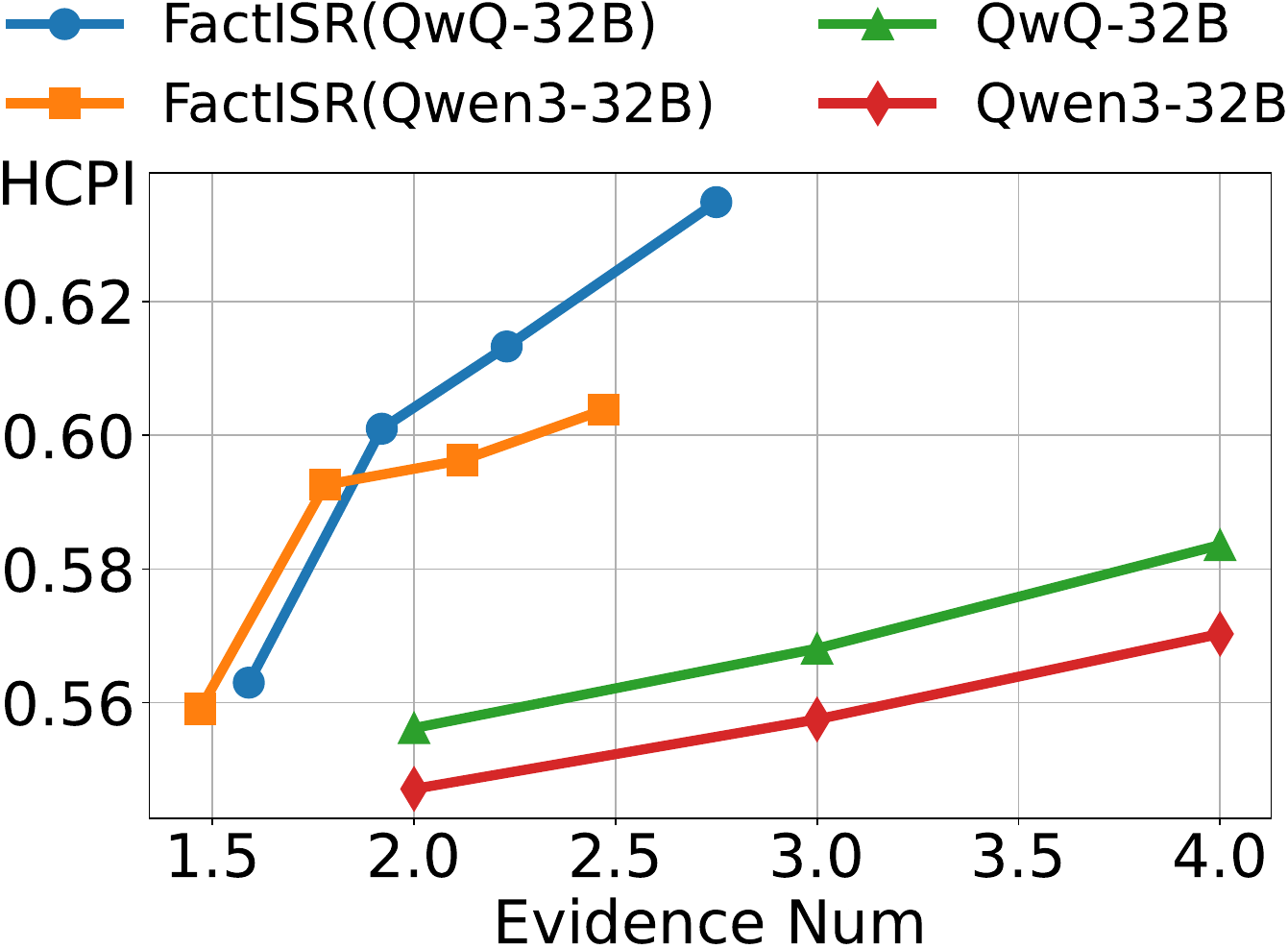}
\includegraphics[width=0.235\textwidth]{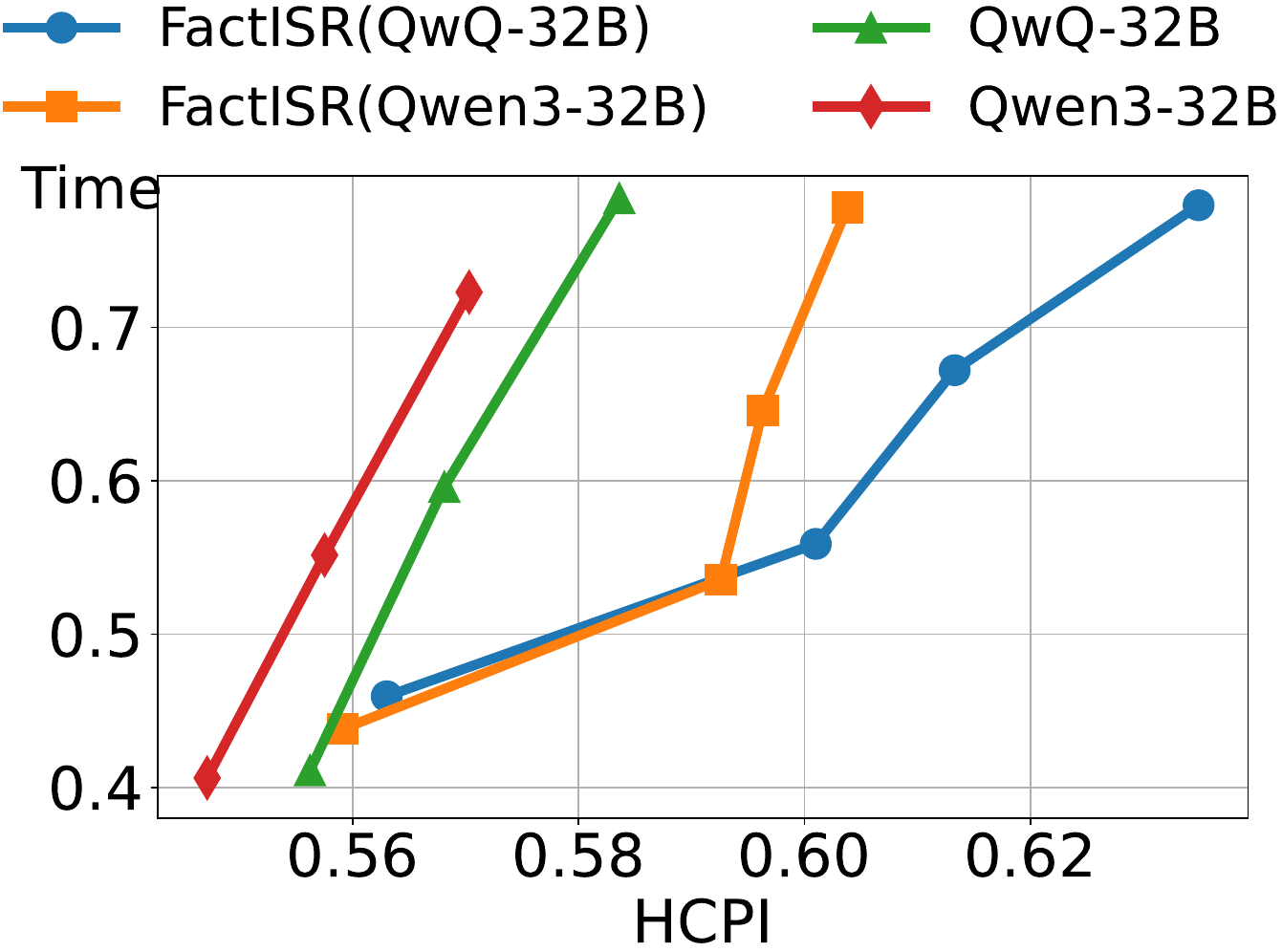}
\caption{Performance comparison between RAG and FactISR under resource constraints.}
\label{fig:Limited_Resources}
\end{figure}

\paragraph{Evaluation with Limited Resources}
We conduct experiments to evaluate the influence of FactISR on HPA performance and checking efficiency of RLMs under resource constraints.


As shown in Figure \ref{fig:Limited_Resources}, we adjust the number of input evidence pieces for RAG and the reflection intensity hyperparameter \( k \) for FactISR. It is evident that FactISR achieves a consistently higher HPA with the same amount of evidence and significantly reduces the average inference time required to reach the same HPA level, thereby greatly improving overall verification efficiency.

\section{Conclusion}
In this paper, we introduce TrendFact, the first benchmark capable of evaluating HPA and all fact-checking tasks. It comprises 7,643 challenging samples with an evidence library containing 366,634 entities through a rigorous construction process. We also propose two novel metrics, ECS and HCPI, to assess the explanation reliability and HPA of automatic fact-checking (AFC) systems. In addition, we present FactISR framework to enhance the HPA and computational efficiency for RLMs-served checking systems. Experimental results demonstrate that TrendFact poses challenges to existing AFC methods, while FactISR effectively improves overall performance.

\section{Limitations}
In this paper, we propose a fact-checking benchmark, TrendFact, which includes structured natural language explanations. However, to improve its real-time relevance, the claims in our dataset are sourced from trending statements on platforms, which require significant human effort to convert into more complex reasoning claims. Additionally, the evidence and explanations in the benchmark are manually gathered and summarized, resulting in high labor costs. We explore whether, in the future, more powerful LLMs with human-like summarization abilities can alleviate this issue. Furthermore, the current benchmark is limited to a single language, and we aim to extend it to a multilingual setting in future work to enhance its broader applicability.

\section{Ethics and Compliance}
The claims in TrendFact are derived from trending platforms and existing datasets, with evidence sourced from public content on the Internet, and its construction does not violate relevant platform rules or legal regulations \cite{calzada2022citizens, chen2021understanding}, making it suitable for academic research. Specifically, the claims collected from trending platforms are taken exclusively from public pages that do not require login. The collection process strictly follows the platform’s open-access agreements (e.g., Weibo Open Platform Agreement: https://open.weibo.com), and only public data is retrieved, without involving any non-public interfaces, private user content, or data requiring authorization. During this process, we only collected information such as trending titles, dates, and urls, which fall within the lowest-risk category, and do not include sensitive data such as user nicknames or user IDs. Date information is retained only to the level of “year–month–day” rather than exact timestamps; additionally, we applied rate limiting ($ \leq 1$ request per second) to avoid imposing load on the servers.

\bibliography{my_ref}

\begin{thebibliography}{32}
\providecommand{\natexlab}[1]{#1}

\bibitem[{qwe(2024)}]{qwen2}
 2024.
\newblock Qwen2 technical report.

\bibitem[{Aly et~al.(2021)Aly, Guo, Schlichtkrull, Thorne, Vlachos, Christodoulopoulos, Cocarascu, and Mittal}]{aly2021feverous}
Rami Aly, Zhijiang Guo, Michael Schlichtkrull, James Thorne, Andreas Vlachos, Christos Christodoulopoulos, Oana Cocarascu, and Arpit Mittal. 2021.
\newblock Feverous: Fact extraction and verification over unstructured and structured information.
\newblock \emph{arXiv preprint arXiv:2106.05707}.

\bibitem[{Aondover et~al.(2024)Aondover, Ebele, Onyejelem, and Akin-Odukoya}]{aondover2024propagation}
Eric~Msughter Aondover, Uchendu~Chinelo Ebele, Timothy~Ekeledirichukwu Onyejelem, and Omolara~Oluwabusayo Akin-Odukoya. 2024.
\newblock Propagation of false information on covid-19 among nigerians on social media.
\newblock \emph{LingLit Journal Scientific Journal for Linguistics and Literature}, 5(3):158--172.

\bibitem[{Atanasova(2024)}]{atanasova2024generating}
Pepa Atanasova. 2024.
\newblock Generating fact checking explanations.
\newblock In \emph{Accountable and Explainable Methods for Complex Reasoning over Text}, pages 83--103. Springer.

\bibitem[{Bilal et~al.(2024)Bilal, Nakov, Procter, and Liakata}]{bilal2024generating}
Iman~Munire Bilal, Preslav Nakov, Rob Procter, and Maria Liakata. 2024.
\newblock Generating unsupervised abstractive explanations for rumour verification.
\newblock \emph{arXiv preprint arXiv:2401.12713}.

\bibitem[{Calzada(2022)}]{calzada2022citizens}
Igor Calzada. 2022.
\newblock Citizens’ data privacy in china: The state of the art of the personal information protection law (pipl).
\newblock \emph{Smart Cities}, 5(3):1129--1150.

\bibitem[{Chen et~al.(2024)Chen, Xiao, Zhang, Luo, Lian, and Liu}]{chen2024bge}
Jianlv Chen, Shitao Xiao, Peitian Zhang, Kun Luo, Defu Lian, and Zheng Liu. 2024.
\newblock Bge m3-embedding: Multi-lingual, multi-functionality, multi-granularity text embeddings through self-knowledge distillation.
\newblock \emph{arXiv preprint arXiv:2402.03216}.

\bibitem[{Chen et~al.(2022)Chen, Sriram, Choi, and Durrett}]{chen2022generating}
Jifan Chen, Aniruddh Sriram, Eunsol Choi, and Greg Durrett. 2022.
\newblock Generating literal and implied subquestions to fact-check complex claims.
\newblock \emph{arXiv preprint arXiv:2205.06938}.

\bibitem[{Chen and Sun(2021)}]{chen2021understanding}
Jihong Chen and Jiabin Sun. 2021.
\newblock Understanding the chinese data security law.
\newblock \emph{International Cybersecurity Law Review}, 2(2):209--221.

\bibitem[{Damasceno et~al.(2024)Damasceno, Rexhepi, Shafer, Whitehouse, Japkowicz, Cavalcante, Corizzo, and Boukouvalas}]{damasceno2024exploiting}
Lucas~P Damasceno, Egzona Rexhepi, Allison Shafer, Ian Whitehouse, Nathalie Japkowicz, Charles~C Cavalcante, Roberto Corizzo, and Zois Boukouvalas. 2024.
\newblock Exploiting sparsity and statistical dependence in multivariate data fusion: an application to misinformation detection for high-impact events.
\newblock \emph{Machine Learning}, 113(4):2183--2205.

\bibitem[{Guo et~al.(2025)Guo, Yang, Zhang, Song, Zhang, Xu, Zhu, Ma, Wang, Bi et~al.}]{guo2025deepseek}
Daya Guo, Dejian Yang, Haowei Zhang, Junxiao Song, Ruoyu Zhang, Runxin Xu, Qihao Zhu, Shirong Ma, Peiyi Wang, Xiao Bi, et~al. 2025.
\newblock Deepseek-r1: Incentivizing reasoning capability in llms via reinforcement learning.
\newblock \emph{arXiv preprint arXiv:2501.12948}.

\bibitem[{Gupta and Srikumar(2021)}]{gupta2021x}
Ashim Gupta and Vivek Srikumar. 2021.
\newblock X-fact: A new benchmark dataset for multilingual fact checking.
\newblock \emph{arXiv preprint arXiv:2106.09248}.

\bibitem[{He et~al.(2023)He, Ahamad, and Kumar}]{he2023reinforcement}
Bing He, Mustaque Ahamad, and Srijan Kumar. 2023.
\newblock Reinforcement learning-based counter-misinformation response generation: a case study of covid-19 vaccine misinformation.
\newblock In \emph{Proceedings of the ACM Web Conference 2023}, pages 2698--2709.

\bibitem[{Hu et~al.(2022)Hu, Guo, Wu, Liu, Wen, and Yu}]{hu2022chef}
Xuming Hu, Zhijiang Guo, GuanYu Wu, Aiwei Liu, Lijie Wen, and Philip~S Yu. 2022.
\newblock Chef: A pilot chinese dataset for evidence-based fact-checking.
\newblock \emph{arXiv preprint arXiv:2206.11863}.

\bibitem[{Hurst et~al.(2024)Hurst, Lerer, Goucher, Perelman, Ramesh, Clark, Ostrow, Welihinda, Hayes, Radford et~al.}]{hurst2024gpt}
Aaron Hurst, Adam Lerer, Adam~P Goucher, Adam Perelman, Aditya Ramesh, Aidan Clark, AJ~Ostrow, Akila Welihinda, Alan Hayes, Alec Radford, et~al. 2024.
\newblock Gpt-4o system card.
\newblock \emph{arXiv preprint arXiv:2410.21276}.

\bibitem[{Jiang et~al.(2020)Jiang, Bordia, Zhong, Dognin, Singh, and Bansal}]{jiang2020hover}
Yichen Jiang, Shikha Bordia, Zheng Zhong, Charles Dognin, Maneesh Singh, and Mohit Bansal. 2020.
\newblock Hover: A dataset for many-hop fact extraction and claim verification.
\newblock \emph{arXiv preprint arXiv:2011.03088}.

\bibitem[{Kao and Yen(2024)}]{kao2024we}
Wei-Yu Kao and An-Zi Yen. 2024.
\newblock How we refute claims: Automatic fact-checking through flaw identification and explanation.
\newblock In \emph{Companion Proceedings of the ACM on Web Conference 2024}, pages 758--761.

\bibitem[{Li et~al.(2025)Li, Jin, Dong, Qian, Zhu, Wu, Wen, and Dou}]{li2025webthinker}
Xiaoxi Li, Jiajie Jin, Guanting Dong, Hongjin Qian, Yutao Zhu, Yongkang Wu, Ji-Rong Wen, and Zhicheng Dou. 2025.
\newblock Webthinker: Empowering large reasoning models with deep research capability.
\newblock \emph{arXiv preprint arXiv:2504.21776}.

\bibitem[{Lin et~al.(2024)Lin, Lin, Yeh, Li, Hu, Hsu, Lee, and Kao}]{lin2024cfever}
Ying-Jia Lin, Chun-Yi Lin, Chia-Jen Yeh, Yi-Ting Li, Yun-Yu Hu, Chih-Hao Hsu, Mei-Feng Lee, and Hung-Yu Kao. 2024.
\newblock Cfever: A chinese fact extraction and verification dataset.
\newblock In \emph{Proceedings of the AAAI Conference on Artificial Intelligence}, volume~38, pages 18626--18634.

\bibitem[{Liu et~al.(2024)Liu, Feng, Xue, Wang, Wu, Lu, Zhao, Deng, Zhang, Ruan et~al.}]{liu2024deepseek}
Aixin Liu, Bei Feng, Bing Xue, Bingxuan Wang, Bochao Wu, Chengda Lu, Chenggang Zhao, Chengqi Deng, Chenyu Zhang, Chong Ruan, et~al. 2024.
\newblock Deepseek-v3 technical report.
\newblock \emph{arXiv preprint arXiv:2412.19437}.

\bibitem[{Pan et~al.(2023)Pan, Wu, Lu, Luu, Wang, Kan, and Nakov}]{pan2023fact}
Liangming Pan, Xiaobao Wu, Xinyuan Lu, Anh~Tuan Luu, William~Yang Wang, Min-Yen Kan, and Preslav Nakov. 2023.
\newblock Fact-checking complex claims with program-guided reasoning.
\newblock \emph{arXiv preprint arXiv:2305.12744}.

\bibitem[{Popat et~al.(2018)Popat, Mukherjee, Yates, and Weikum}]{popat2018declare}
Kashyap Popat, Subhabrata Mukherjee, Andrew Yates, and Gerhard Weikum. 2018.
\newblock Declare: Debunking fake news and false claims using evidence-aware deep learning.
\newblock \emph{arXiv preprint arXiv:1809.06416}.

\bibitem[{Rani et~al.(2023)Rani, Tonmoy, Dalal, Gautam, Chakraborty, Chadha, Sheth, and Das}]{rani2023factify}
Anku Rani, SM~Tonmoy, Dwip Dalal, Shreya Gautam, Megha Chakraborty, Aman Chadha, Amit Sheth, and Amitava Das. 2023.
\newblock Factify-5wqa: 5w aspect-based fact verification through question answering.
\newblock \emph{arXiv preprint arXiv:2305.04329}.

\bibitem[{Schlichtkrull et~al.(2024)Schlichtkrull, Guo, and Vlachos}]{schlichtkrull2024averitec}
Michael Schlichtkrull, Zhijiang Guo, and Andreas Vlachos. 2024.
\newblock Averitec: A dataset for real-world claim verification with evidence from the web.
\newblock \emph{Advances in Neural Information Processing Systems}, 36.

\bibitem[{Sehat et~al.(2024)Sehat, Li, Nie, Prabhakar, and Zhang}]{sehat2024misinformation}
Connie~Moon Sehat, Ryan Li, Peipei Nie, Tarunima Prabhakar, and Amy~X Zhang. 2024.
\newblock Misinformation as a harm: structured approaches for fact-checking prioritization.
\newblock \emph{Proceedings of the ACM on human-computer interaction}, 8(CSCW1):1--36.

\bibitem[{Solovev and Pr{\"o}llochs(2022)}]{solovev2022moral}
Kirill Solovev and Nicolas Pr{\"o}llochs. 2022.
\newblock Moral emotions shape the virality of covid-19 misinformation on social media.
\newblock In \emph{Proceedings of the ACM web conference 2022}, pages 3706--3717.

\bibitem[{Thorne and Vlachos(2017)}]{thorne2017extensible}
James Thorne and Andreas Vlachos. 2017.
\newblock An extensible framework for verification of numerical claims.
\newblock In \emph{Proceedings of the Software Demonstrations of the 15th Conference of the European Chapter of the Association for Computational Linguistics}, pages 37--40. Association for Computational Linguistics.

\bibitem[{Thorne et~al.(2018)Thorne, Vlachos, Christodoulopoulos, and Mittal}]{thorne2018fever}
James Thorne, Andreas Vlachos, Christos Christodoulopoulos, and Arpit Mittal. 2018.
\newblock Fever: a large-scale dataset for fact extraction and verification.
\newblock \emph{arXiv preprint arXiv:1803.05355}.

\bibitem[{van Der~Linden et~al.(2020)van Der~Linden, Roozenbeek, and Compton}]{van2020inoculating}
Sander van Der~Linden, Jon Roozenbeek, and Josh Compton. 2020.
\newblock Inoculating against fake news about covid-19.
\newblock \emph{Frontiers in psychology}, page 2928.

\bibitem[{Venktesh et~al.(2024)Venktesh, Anand, Anand, and Setty}]{venktesh2024quantemp}
V~Venktesh, Abhijit Anand, Avishek Anand, and Vinay Setty. 2024.
\newblock Quantemp: A real-world open-domain benchmark for fact-checking numerical claims.
\newblock \emph{arXiv preprint arxiv:2403.17169}.

\bibitem[{Wang and Shu(2023)}]{wang2023explainable}
H~Wang and K~Shu. 2023.
\newblock Explainable claim verification via knowledge-grounded reasoning with large language models. arxiv preprint arxiv: 231005253.

\bibitem[{Yang et~al.(2025)Yang, Li, Yang, Zhang, Hui, Zheng, Yu, Gao, Huang, Lv et~al.}]{yang2025qwen3}
An~Yang, Anfeng Li, Baosong Yang, Beichen Zhang, Binyuan Hui, Bo~Zheng, Bowen Yu, Chang Gao, Chengen Huang, Chenxu Lv, et~al. 2025.
\newblock Qwen3 technical report.
\newblock \emph{arXiv preprint arXiv:2505.09388}.

\end{thebibliography}
\appendix

\section{Details of Data Attributes}
\label{app:detial_data_attributes}
\begin{figure}[!ht]
\centering
\includegraphics[width=\linewidth]{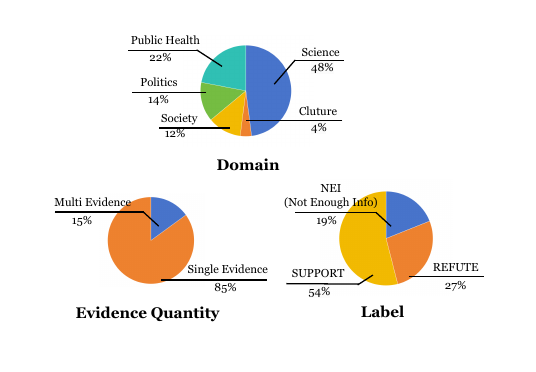} 
\caption{Overview of the data distribution, including labels, gold evidence count, and domains.}
\label{fig:distribution}
\end{figure}

The hotspot indicators of the TrendFact sample include: views, discussions, posts, and engagements. These indicators are crucial for assessing the hotspot perception capabilities of fact-checking systems. Specifically, views represent the number of times the sample has been viewed on sampled trending platforms; discussions indicate the number of times the sample has been discussed; posts refer to the number of posts triggered by the sample; engagements represent the number of users involved. Additionally, the influence score, which is assessed by an LLM to indicate the potential threat level if the claim were false, ranges from 1 to 5, with 5 being the highest threat. This score is also a key component in evaluating the hotspot perception capabilities of fact-checking systems. Figure \ref{fig:distribution} shows the data distribution of TrendFact samples, including labels, gold evidence count, and domains.

\begin{figure*}[!ht]
\centering
\includegraphics[width=1\linewidth]{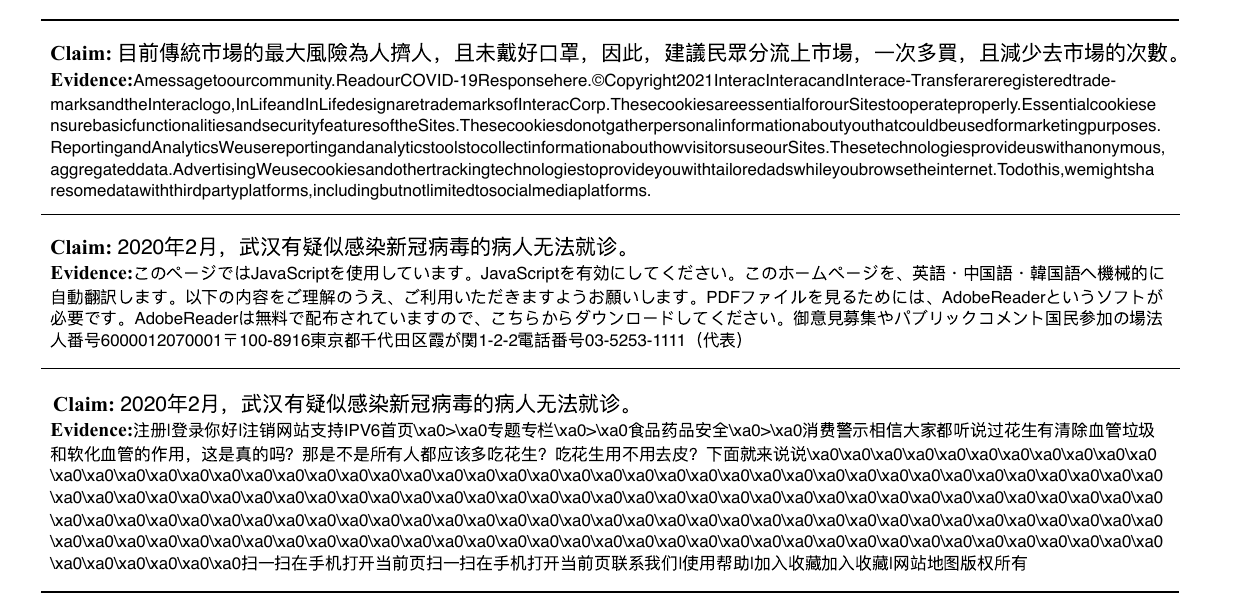} 
\caption{Examples of CHEF Dataset Cleaning.}
\label{chef_clean_example}
\end{figure*}

\begin{table}[t]
\centering
\small 
\begin{tabular}{llclll}
\toprule
Category & F-D & F-C   & T-CD  & T-PC  & T-FC  \\ \midrule
Acc      & 100 & 98.96 & 96.82 & 95.68 & 96.34 \\ \bottomrule
\end{tabular}
\caption{Human-LLM agreement rates (\%) across five ECS categories, based on manual expert validation of over 100 randomly sampled instances.}
\label{tab:LaaJ_eva}
\end{table}

\begin{figure*}[!ht]
\centering
\includegraphics[width=1\linewidth]{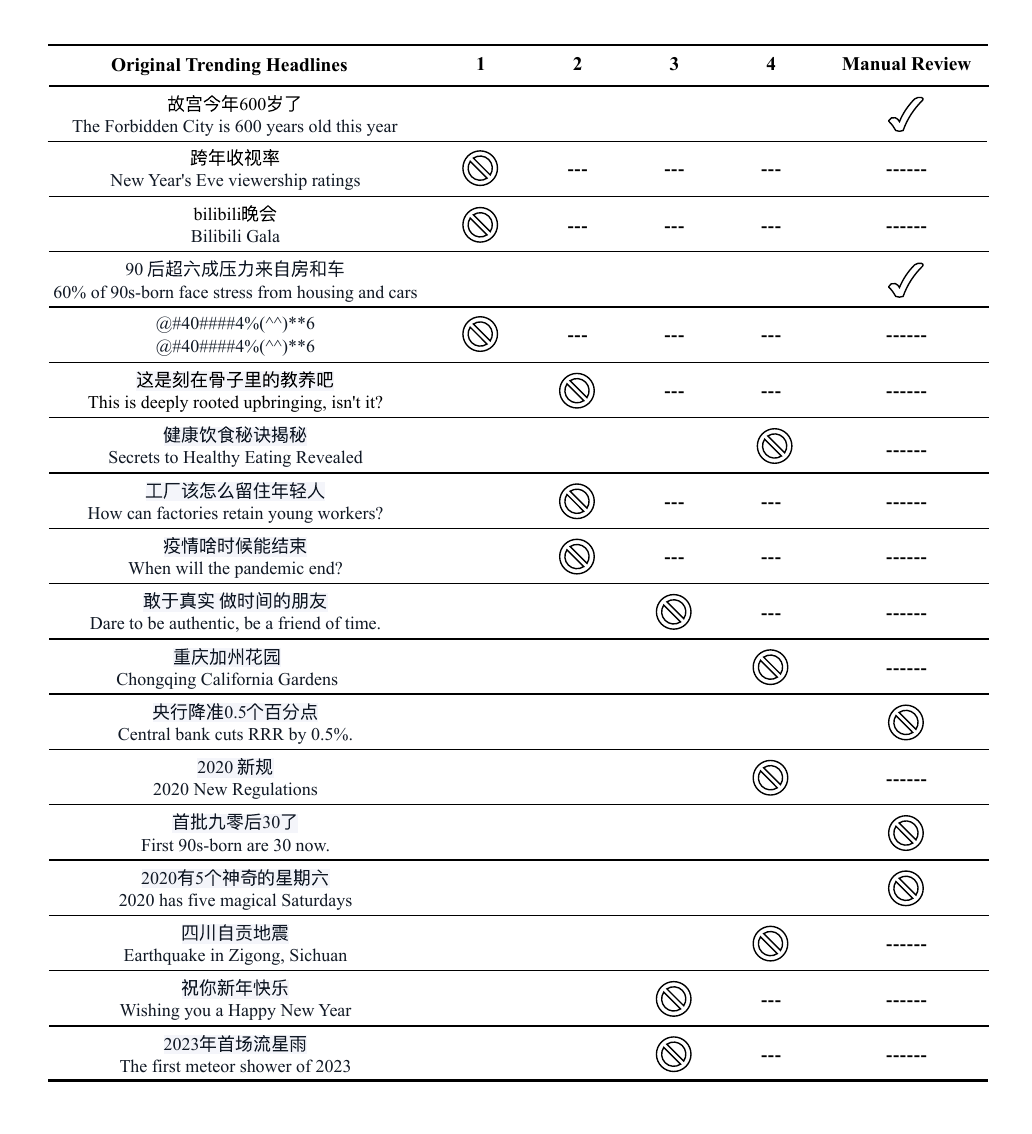} 
\caption{Examples of Progressively Staged Data Filtering Workflow for Fact-Checking Potential Data Selection.}
\label{filtering_example}
\end{figure*}

\begin{table*}[h]
\small 
\centering
\renewcommand{\arraystretch}{1.5} 
\begin{tabularx}{0.98\textwidth}{c|X}
\toprule
\textbf{Iteration} & \textbf{Prompt}                                       \\ \hline
\makecell{1}         & You are a trending topics analysis assistant, capable of accurately identifying the category of trending topics, mainly referring to trends on platforms like Weibo and Baidu.

I will provide you with data on trending topics, and you need to help me determine their category.

Note: I do not want entertainment-related trending topics. This means you do not need to output specific categories; you only need to decide whether a trending topic belongs to the entertainment category, and simply output one word: "Yes" or "No."

Next, I will give you several examples for your reference in making judgments and outputs.

\textbf{\{Example Trending Topics\}}

Note: To emphasize again, you need to determine if a trending topic belongs to the entertainment category, and output only one word (Yes/No)!

Note: If the trending topic is garbled text, also output No! \\ \midrule
2         & You are a trending topics analysis assistant, capable of accurately analyzing the category of trending topics, mainly referring to trends such as Weibo and Baidu hot searches.

I will provide you with some trending topics data, and you need to help me determine whether these data are in question form.

Note, you do not need to output specific categories; you only need to determine whether a trending topic is in question form and simply output one word: "Yes" or "No."

Next, I will give you several examples for your reference to make judgments and outputs.

\textbf{\{Example Trending Topics\}}

Note: To emphasize again, you need to determine if the trending topic is in the form of a question and output only one word (Yes/No)!

Note: Questions here may not necessarily contain a question mark or have obvious question features; they might be guiding sentences designed to attract clicks.                                             \\ \midrule
3         & You are a fact-checking assistant, capable of accurately determining whether the current input can serve as a sample for fact-checking.

It is known that a fact-checking task involves assessing the truthfulness of a claim based on provided evidence.

However, I do not need to assess its truthfulness now; rather, I want to determine whether the current input has the potential to serve as a sample for a fact-checking dataset.

I will provide you with real trending topics data from Weibo, and you need to help me assess whether these data have the potential to be included as samples in a fact-checking dataset.

Note, you do not need to identify where the potential lies; you only need to output one word: "Yes" or "No."

Next, I will give you several examples for your reference to make judgments and outputs.
Examples are as follows:

\textbf{\{Example Trending Topics\}}

Note: You need to assess whether the trending topic has the potential to serve as a sample for a fact-checking dataset, and output only one word (Yes/No)!

Note: Having potential means it contains elements that can be assessed and requires support from evidence, rather than abrupt statements or blessing words, etc.!                                             \\ \midrule
4         & You are a fact-checking assistant, capable of accurately determining whether the current input can serve as a sample for fact-checking.

It is known that a fact-checking task involves assessing the truthfulness of a claim based on provided evidence.

However, I do not need to assess its truthfulness now; rather, I want to determine whether the current input has the potential to serve as a sample for fact-checking.

More specifically, if the current input is merely in noun form, then it does not have the potential to be included as a sample in a fact-checking dataset.

I will provide you with real trending topics data from Weibo, and you need to help me assess whether these data have the potential to be included as samples in a fact-checking dataset.

Note, you do not need to identify where the potential lies; you only need to output one word: "Yes" or "No."

Next, I will give you several examples for your reference to make judgments and outputs.
Examples are as follows:

\textbf{\{More Challenging Trending Topic Examples\}}

Note: To emphasize again, you need to assess whether the trending topic has the potential to serve as a sample for a fact-checking dataset, and output only one word (Yes/No)!

Note: Having potential means it is not merely a noun and contains elements that can be assessed, requiring support from evidence, rather than abrupt statements or blessing words, etc.!                                             \\ \bottomrule
\end{tabularx}
\caption{Progressively Staged Prompts for Fact-Checking Potential Selection.}
\label{filtering_prompt}
\end{table*}

\section{Data Cleaning and Hard Example Selection}
\label{app:noise_data}
Figure \ref{chef_clean_example} demonstrates data cleaning examples from the CHEF dataset, primarily showing the removal of samples containing extensive garbled text in evidence sources. For filtering trending headlines with fact-checking potential from social platforms, this paper implements a progressive human-AI collaborative filtering strategy. The pipeline sequentially eliminates headlines at different stages: (1) Initial filtering using large language models (LLMs) with stage-specific prompts, followed by (2) manual verification when sample quantities become manageable. This multi-stage approach yields challenging yet verifiable candidate samples through layered refinement.
Table \ref{filtering_prompt} and Figure \ref{filtering_example} respectively present examples of stage-specific prompts and the sample filtering workflow.

\begin{table*}[!htbp]
\centering
\begin{tabular}{
    >{\raggedright\arraybackslash}p{3.5cm}
    >{\arraybackslash}p{5.5cm}
    >{\arraybackslash}p{5cm}
}
\toprule
\textbf{Factor Category} & \textbf{Definition} & \textbf{Rewriting Mechanism} \\
\midrule
Temporal Anchoring & Adding/specifying temporal reference & Transforming vague temporal expressions into specific time nodes \\ \midrule

Data Granularity & Disaggregating composite data into verifiable units & Decomposing aggregated data into independently verifiable dimensions \\ \midrule

Ambiguity Resolution & Eliminating probabilistic/uncertain expressions & Replacing fuzzy quantifiers with deterministic statements \\ \midrule

Comparative Standard & Establishing quantifiable reference standards & Introducing quantified comparison objects and proportions \\ \midrule

Domain Knowledge & Incorporating professional contextual information & Supplementing industry-specific parameters or mechanisms \\ \midrule

Source Implication & Indirectly indicating information provenance & Using industry-characteristic expressions to imply data sources \\
\bottomrule
\end{tabular}
\caption{Fact-Checking Claim Rewriting Factor}
\label{tab:en_factors}
\end{table*}

\begin{figure*}[!ht]
\centering
\includegraphics[width=1\linewidth]{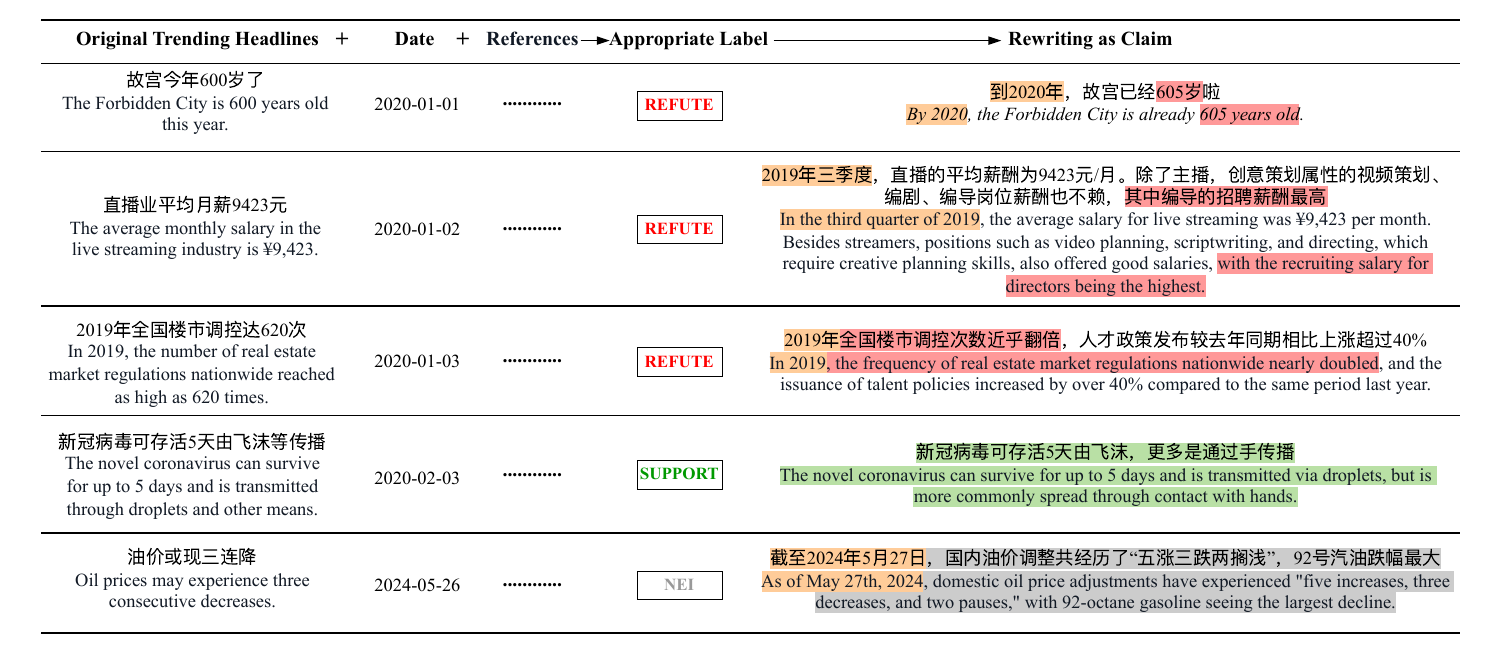} 
\caption{Examples of Rewriting Trending Headlines into Fact-Checkable Claims.}
\label{rewriting_example}
\end{figure*}

\section{Fact-Checking Claim Rewriting Factor}
\label{app:rewrite_factor}
As shown in Table \ref{tab:en_factors}, six critical factors are proposed to systematically transform social media headlines into verifiable claims: (1) Temporal Anchoring converts vague temporal expressions into specific time references; (2) Data Granularity decomposes aggregated data into measurable units; (3) Ambiguity Resolution replaces probabilistic terms with deterministic statements; (4) Comparative Standard introduces quantifiable benchmarks; (5) Domain Knowledge integrates industry-specific parameters; and (6) Source Implication embeds provenance cues. These factors collectively enhance claim verifiability while preserving semantic coherence for automated processing.
Concrete rewriting examples are illustrated in Figure \ref{rewriting_example}.


\section{Annotate Evaluation Criteria}
\label{app:human_evaluation_criteria}
Our human evaluation criteria, as illustrated in Figure \ref{framework}, are divided into three components: attribute-level review, fact verification review, and explanation generation review. Specifically, in the attribute-level review, we assess the individual quality and mutual consistency of each sample's claim, evidence, and explanation. For the fact verification review, annotators independently determine the veracity of each claim based on the corresponding evidence and compare their judgments with the labeled result. In the explanation generation review, we manually verify whether the annotated explanation meets the predefined standards, given the claim, evidence, and label of the sample.

\begin{figure*}[!t]
\centering
\includegraphics[width=0.85\linewidth]{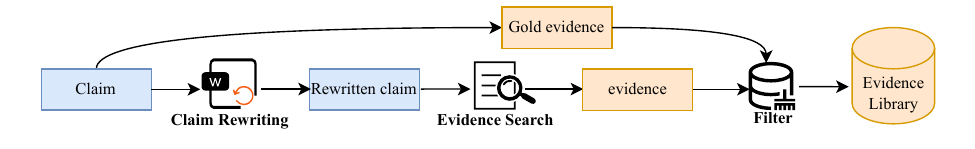} 
\caption{Evidence Library Construction Process.}
\label{Evidence_Library_construction_process}
\end{figure*}

\begin{figure*}[!b]
\centering
\includegraphics[width=1\linewidth]{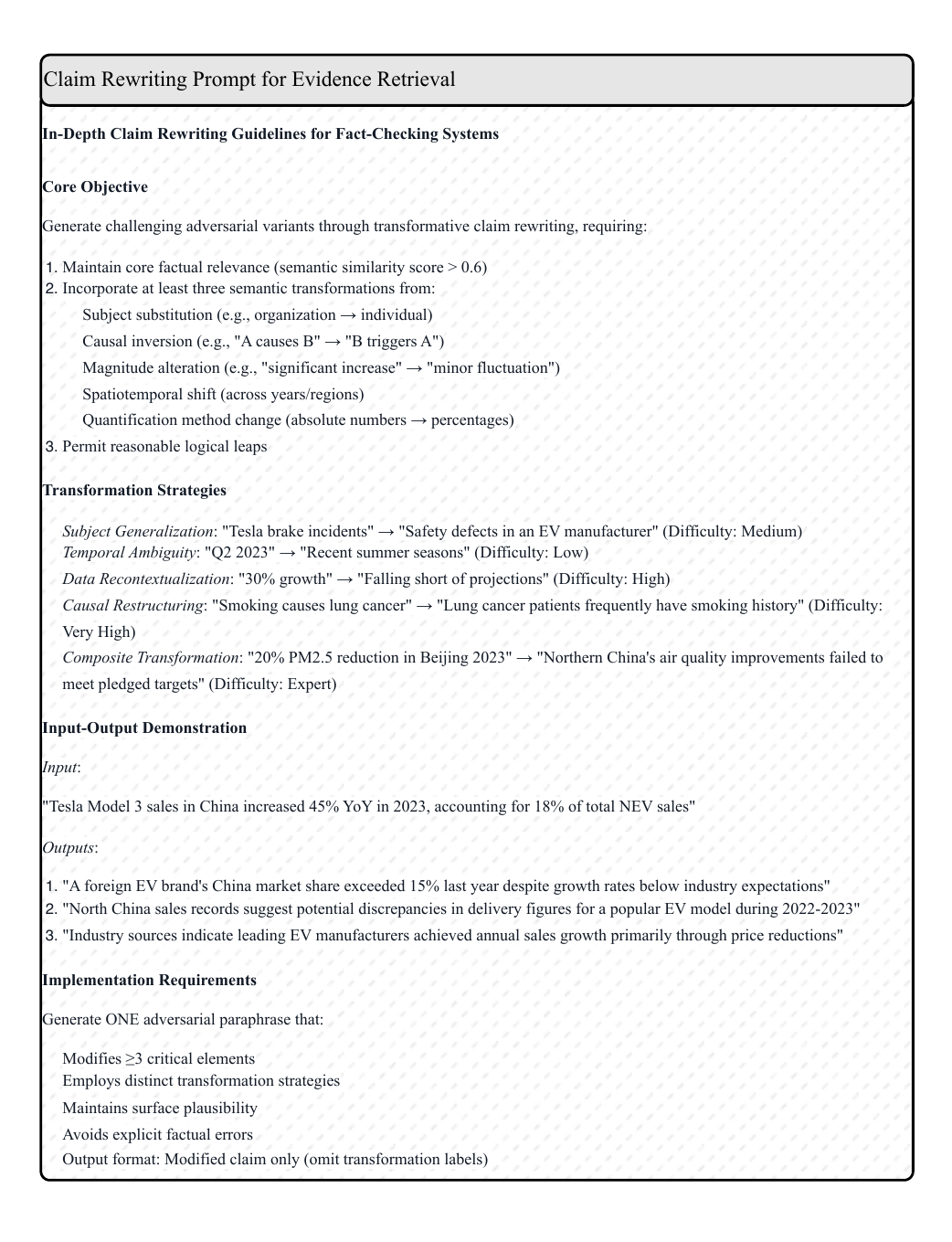} 
\caption{Claim Rewriting Prompt for Evidence Retrieval.}
\label{Claim_Rewriting_Prompt_for_Evidence_Retrieval}
\end{figure*}


\section{Evidence Library Construction Process}
\label{ELC:more_details}
First, we employ GPT-4.1 to extract multiple key-element phrases from the original claims and to generate rewritten statements based on the original claims. Figure \ref{Claim_Rewriting_Prompt_for_Evidence_Retrieval} showcases the specific rewriting prompt. Specifically, we transform original statements into challenging variants by maintaining core fact relevance (relevance score >0.6) and introducing at least three types of semantic alterations: subject displacement (e.g., institution → individual), causal inversion (e.g., "A leads to B" → "B triggers A"), degree conversion (e.g., "significant growth" → "slight fluctuation"), spatiotemporal shift (across years/regions), and quantification method change (absolute value → percentage), allowing for reasonable logical leaps. The rewritten claims are then processed through Bing web retrieval, retaining the top 10 search results and extracting their textual content. Subsequently, we merge and deduplicate the newly collected evidence with pre-existing gold evidence, followed by publication date crawling for each webpage to finalize the evidence library.


\section{Human Validation of LLM-as-a-Judge for ECS}
\label{sec:LaaJ}
Table \ref{tab:LaaJ_eva} demonstrates that the LLM-as-a-Judge evaluator achieves strong agreement with human expert assessments across all ECS categories.

\section{Prompt of FactISR}
\begin{figure*}[!ht]
\centering
\includegraphics[width=1\linewidth]{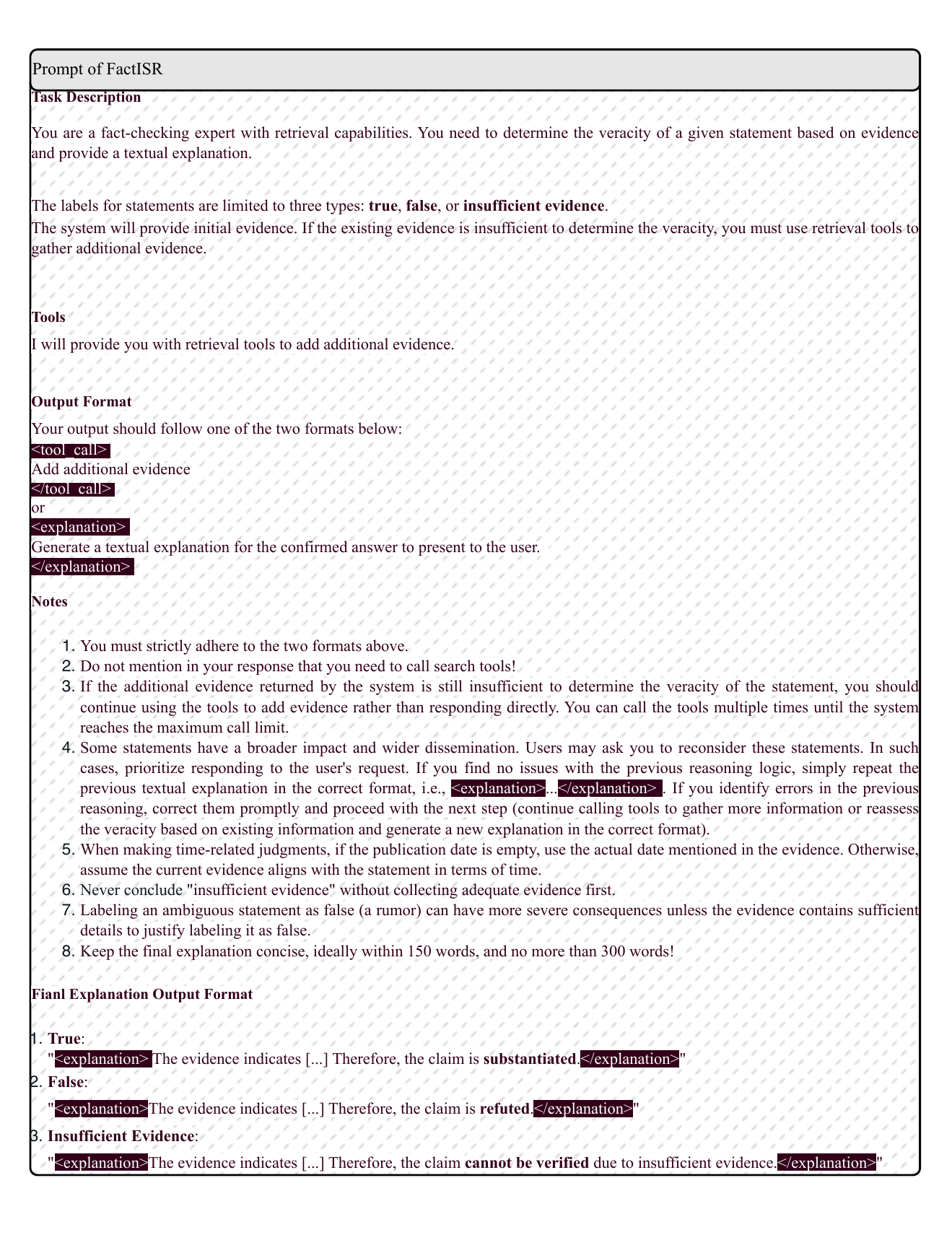} 
\caption{English-Translated FactISR Prompt (partial formatting directives omitted for space constraints).}
\label{prompt_template}
\end{figure*}

Figure \ref{prompt_template} shows the prompt of FactISR.

\begin{figure*}[!ht]
\centering
\includegraphics[width=1\linewidth]{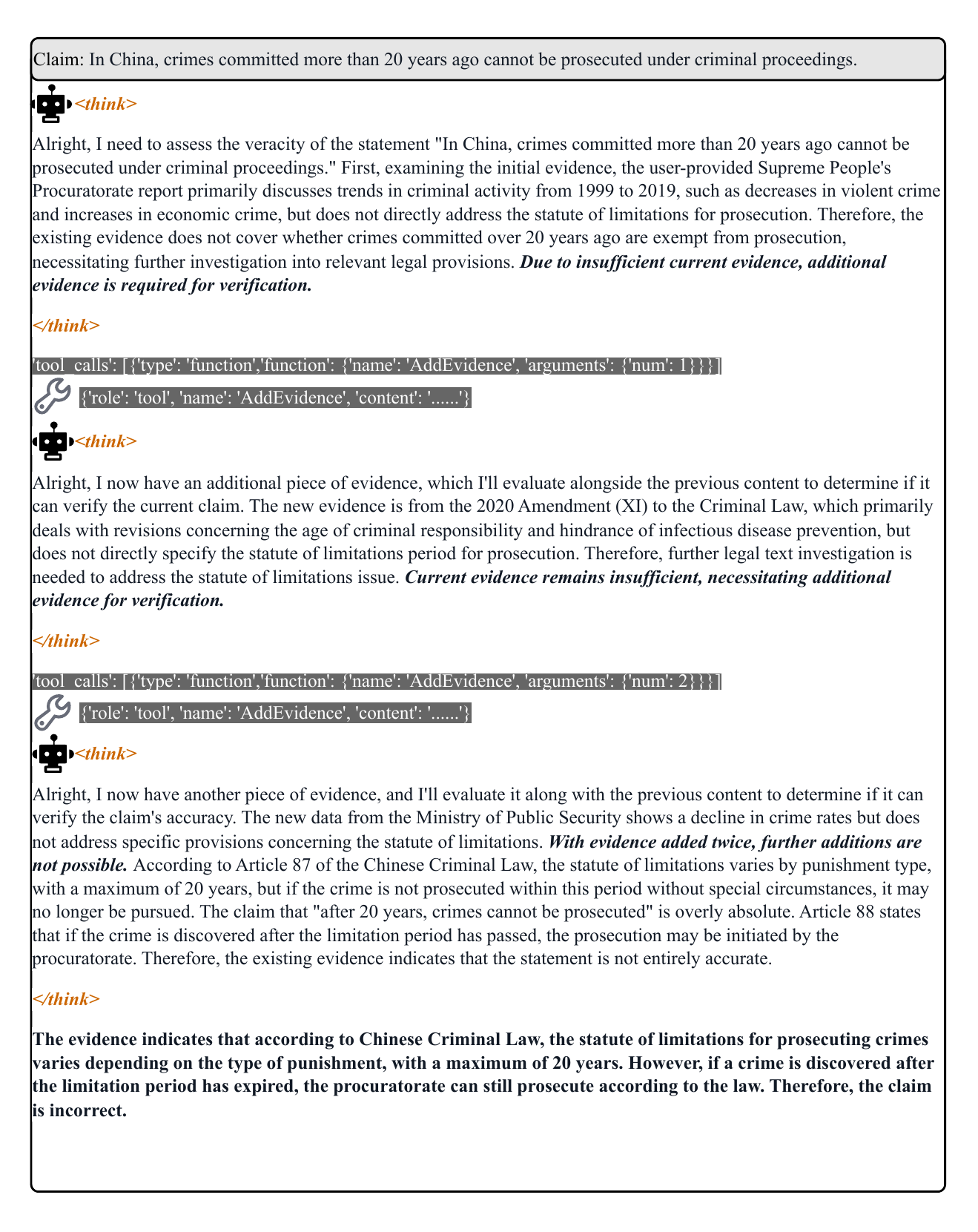} 
\caption{Example of Dynamic Evidence Addition.}
\label{DEA_example}
\end{figure*}

\section{Example of FactISR}
\begin{figure*}[!ht]
\centering
\includegraphics[width=1\linewidth]{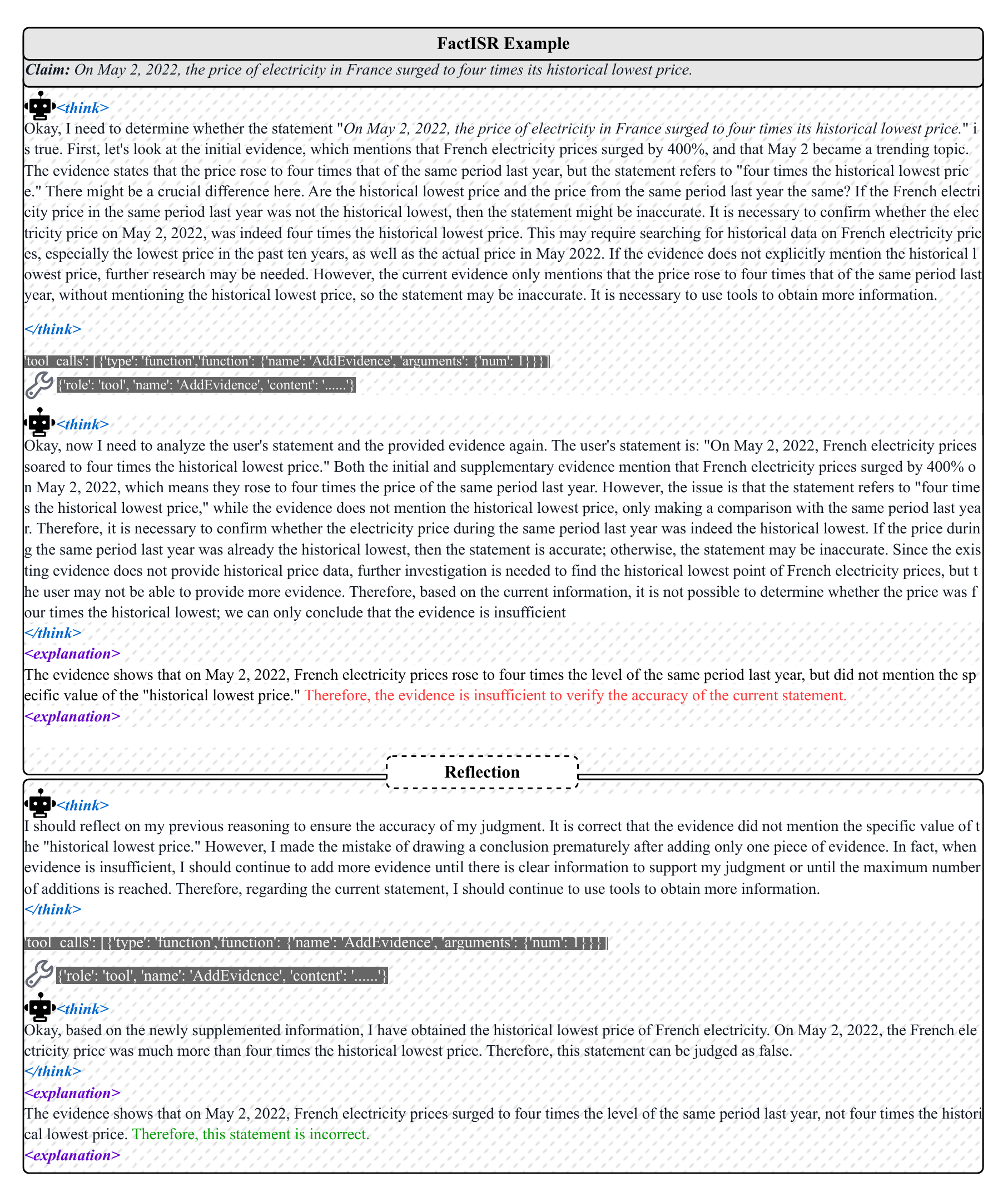} 
\caption{An Example of FactISR.}
\label{example_rd}
\end{figure*}

Figure \ref{example_rd} illustrates an example of FactISR. Without ISR, the model directly outputs a conclusion of insufficient evidence and prematurely ends the reasoning process. Our reflection mechanism encourages the model to reassess its previous judgment, leading to a reconsideration that ultimately results in the correct outcome.

\section{Experiments Under Gold Evidence Conditions}
Tables \ref{tab:gold_evi_compare_baseline_fact_verification} and \ref{tab:gold_evi_compare_baseline_explanation_generation} present experimental results of fact verification and explanation generation tasks under gold evidence conditions for LLMs and RLMs. Since gold evidence was pre-defined (rendering the DEA module inapplicable), our FactISR method is excluded from this comparison. The results demonstrate significant improvements in fact verification metrics (accuracy: +5-10 percentage points; F1) and explanation generation quality. Specifically, the fact verification accuracy of these methods significantly improved by 5 to 10 percentage points, with DeepSeek-R1 and o1-preview achieving scores of 77.92\% and 78.98\%, respectively. Similarly, for the explanation generation task, DeepSeek-V3 achieved a BLEU-4 score of 0.3573, which is nearly 0.1 points higher than when using retrieval-based evidence.

\section{Details of the Experimental Settings}
\label{Experimental_Settings:more_details}
We conduct experiments on PyTorch\footnote{\url{https://pytorch.org/}} and $2\times A100$ GPUs. The evaluation for ECS was conducted using GPT-4.1, while BERTScore evaluations are conducted on \textit{chinese-bert-Base}\footnote{\url{https://huggingface.co/google-bert/bert-base-chinese}}.
The small constant \( \epsilon \) to prevent division by zero, the hyperparameter \( k \)  controlling the reflection strength, and the decay factor \(\gamma\) are set to  \( 1 \times 10^{-8} \), 1 and 0.5, respectively.
The maximum number of reflections is set to 3.
The maximum input length is set to 16k, while the maximum output lengths for LLMs and RLMs are set to 300 and 5k.
The maximum length of retrieved results is 3k.
The maximum number of retrievals is set to 3, and to ensure fair comparison, the maximum number of dynamically added evidences by our DEA is also limited to the same.
All inference experiments utilized greedy search as the strategy.
In this paper, for the HCPI metric, the values of $\alpha$, $\beta$, $\kappa$ and $\lambda$ used to calculate the influence score are set to 0.05, 0.2, 0.15, and 0.6, respectively. Missing values are imputed using the 25th percentile, and the scores are scaled to ensure that the ratio of the maximum to the minimum influence score remains within a factor of 10.







\begin{table*}[!b]
\centering
\renewcommand{\arraystretch}{1.2} 
\begin{tabular}{l|cccc}
\toprule
Methods              & Acc   & F1    & P & R \\ \midrule
PROGRAM-FC           & 56.55 & 54.05 & 54.17     & 56.62  \\
CLAIMDECOMP          & 59.35 & 56.86 & 56.65     & 59.41  \\ \midrule
Qwen-72B-instruct    & 65.14 & 60.56 & 66.97     & 63.65  \\ 
QwQ-32B-Preview      & 65.31 & 61.76 & 63.68     & 65.53  \\
DeepSeek-V3          & 63.74 & 60.31 & 66.09     & 63.96  \\
GPT-4.1              & 72.29 & 69.68 & 69.02     & 72.88  \\
DeepSeek-R1          & 77.92 & 72.56 & 73.72     & 72.64  \\
o1-preview           & \textbf{78.98} & \textbf{75.16} & \textbf{75.13}     & \textbf{75.72}  \\ \bottomrule
\end{tabular}
\caption{Experimental Results of Baselines Under Gold Evidence Conditions in Fact Verification Task.}
\label{tab:gold_evi_compare_baseline_fact_verification}
\end{table*}

\begin{table*}[!t]
\centering
\renewcommand{\arraystretch}{1.2} 
\begin{tabular}{l|cccccc}
\toprule
Methods              & BLEU-4 & BERTScore & ROUGE-1 & ROUGE-2 & ROUGE-L & ECS \\ \midrule
QwQ-32B-Preview      & 0.2093 & 0.7804    & 0.5330  & 0.3459  & 0.4669  & 0.8198 \\
Qwen-72B-instruct    &  0.3366  & 0.8364     & 0.6441  &  0.4589  & 0.5906  & 0.7787 \\ 
DeepSeek-V3          & \textbf{0.3573} & \textbf{0.8432}    & \textbf{0.6596}  & \textbf{0.4805}  & \textbf{0.6087}  & 0.7812 \\
GPT-4.1              & 0.2958 & 0.8270    & 0.6191  & 0.4189  & 0.5561  & 0.8622 \\
DeepSeek-R1          & 0.2705 & 0.8143    & 0.5832  & 0.3821  & 0.5188  & \textbf{0.9115} \\
o1-preview           & 0.2693 & 0.8022    & 0.5602  & 0.3960  & 0.5206  & 0.8986 \\ \bottomrule
\end{tabular}
\caption{Experimental Results of Baselines Under Gold Evidence Conditions in Explanation Generation Task.}
\label{tab:gold_evi_compare_baseline_explanation_generation}
\end{table*}



\end{document}